\documentclass{article}

\usepackage{arxiv}

\usepackage[utf8]{inputenc} 
\usepackage[T1]{fontenc}    
\usepackage{hyperref}       
\usepackage{url}            
\usepackage{booktabs}       
\usepackage{amsfonts}       
\usepackage{nicefrac}       
\usepackage{microtype}      
\usepackage{xcolor}         
\usepackage{caption}
\usepackage{subcaption}
\usepackage{lipsum}
\usepackage{graphicx}
\usepackage{float}
\graphicspath{ {./images/} }

\title{Real-time Vision-based Navigation for a Robot in an Indoor Environment}

\author{
 Sagar Manglani \\
  The Department of Computer Science (CS)\\
  Stanford University\\
  Stanford, CA 94305 \\
  \texttt{sagarm@stanford.edu} \\
  \texttt{manglanisagar@gmail.com} \\
}

\begin{document}
\maketitle
\begin{abstract}
This paper presents a study on the development of an obstacle-avoidance navigation system for autonomous navigation in home environments. The system utilizes vision-based techniques and advanced path-planning algorithms to enable the robot to navigate toward the destination while avoiding obstacles. The performance of the system is evaluated through qualitative and quantitative metrics, highlighting its strengths and limitations. The findings contribute to the advancement of indoor robot navigation, showcasing the potential of vision-based techniques for real-time, autonomous navigation.
\end{abstract}


\section{Introduction}

The objective of this project is to develop a robust obstacle-avoidance navigation system for a low-cost, 3D-printed, four-legged walking robot in home environments. The robot aims to autonomously navigate towards a specified destination point in the lowest amount of time while avoiding obstacles. 

\begin{figure}[H]
     \centering
     \begin{subfigure}[b]{0.4\textwidth}
         \centering
         \includegraphics[width=\textwidth, trim={20cm 0 0 0},clip]{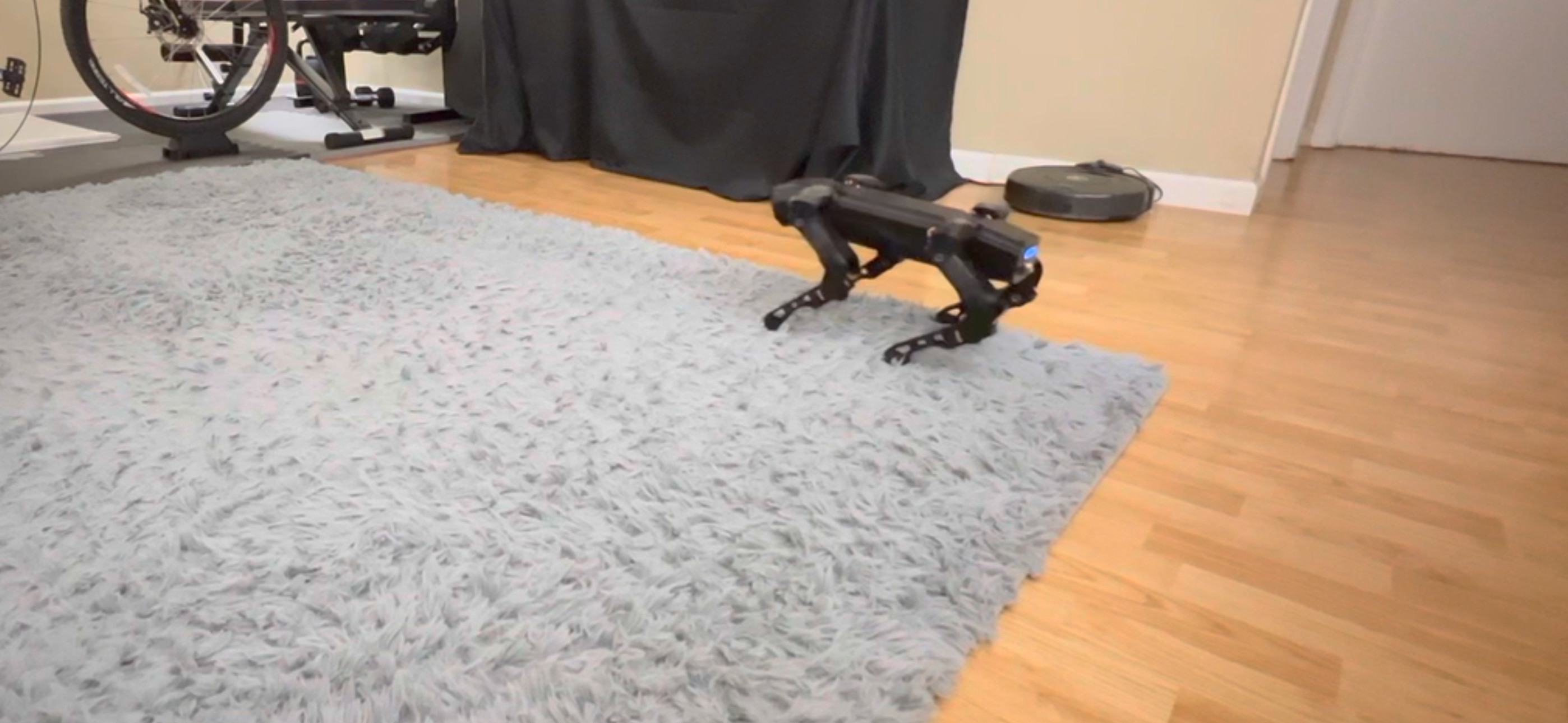}
         \caption{The robot in the environment}
         \label{fig:robot}
     \end{subfigure}
     \hfill
     \begin{subfigure}[b]{0.3\textwidth}
         \centering
         \includegraphics[width=\textwidth]{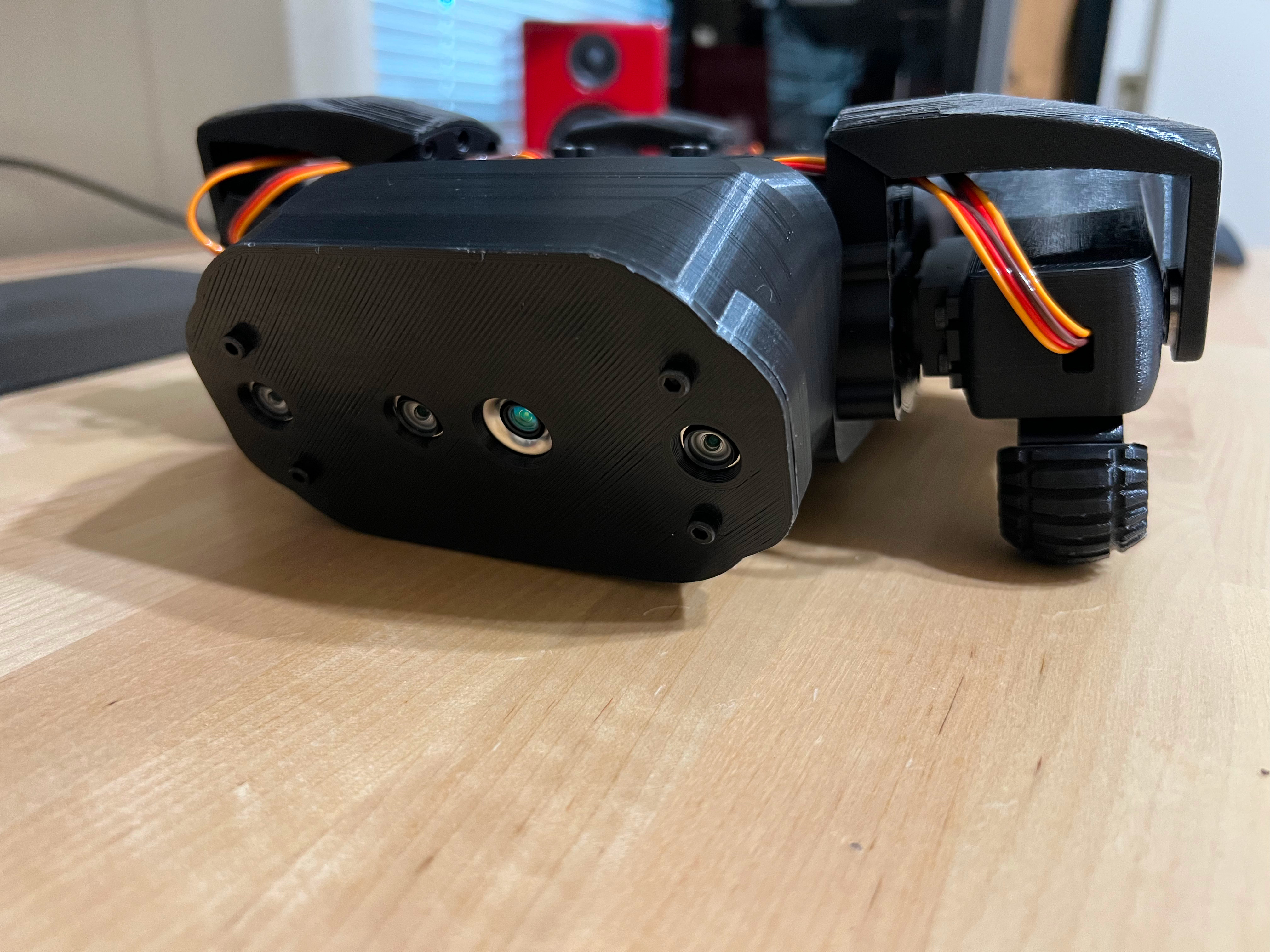}
         \caption{The robot's camera sensor}
         \label{fig:camera}
     \end{subfigure}
     \hfill
     \begin{subfigure}[b]{0.2\textwidth}
         \centering
         \includegraphics[width=\textwidth, trim={0 5cm 0 5cm},clip]{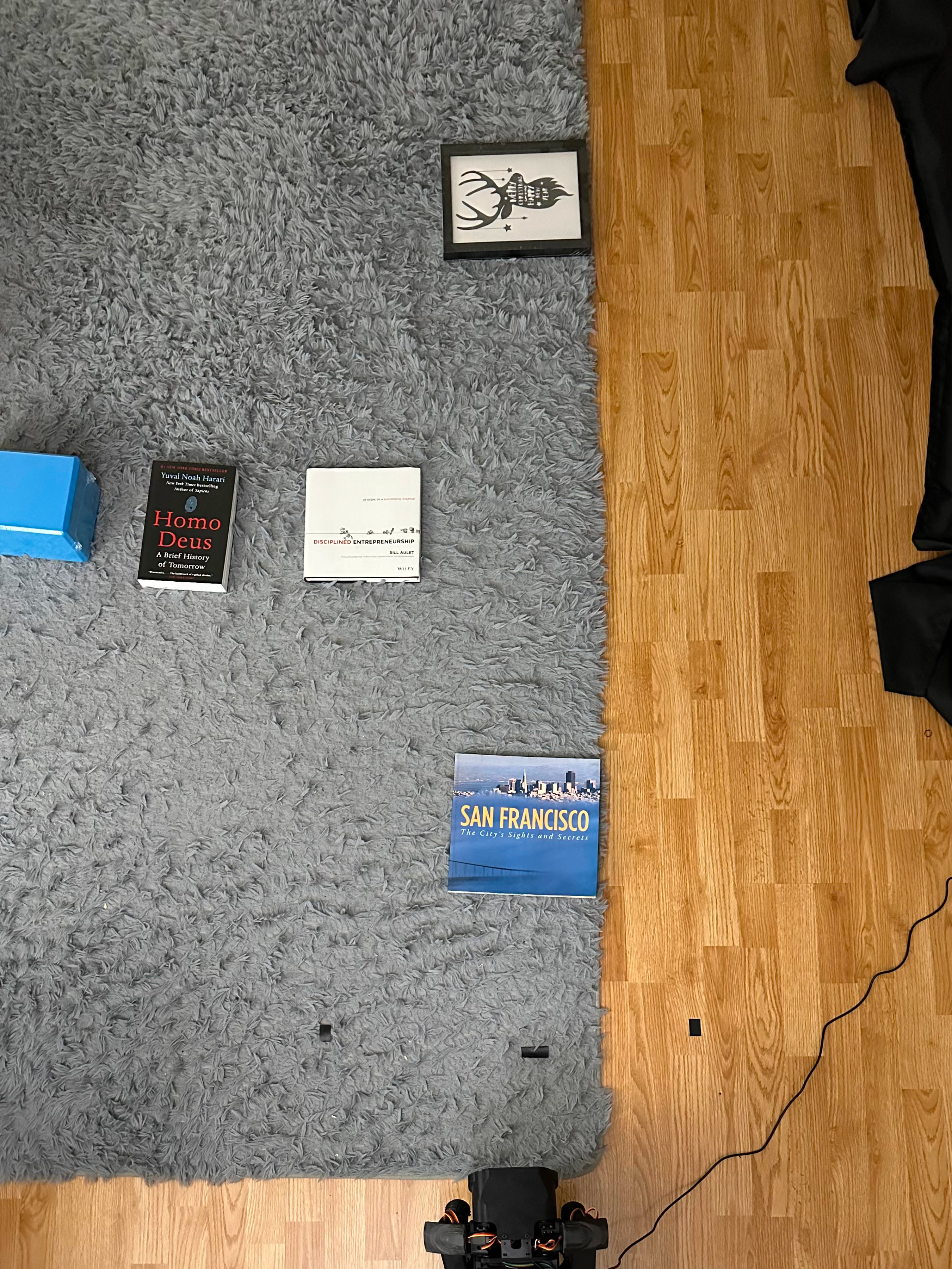}
         \caption{The environment}
         \label{fig:environment}
     \end{subfigure}
        \caption{The setup}
        \label{fig:setup}
\end{figure}

Figure \ref{fig:setup} depicts the fundamental components employed in this project, comprising a robot equipped with an RGBD camera for perception and an Nvidia Jetson Xavier NX for onboard computation. Notably, the navigation system solely relies on visual information by utilizing RGB images exclusively to navigate through the environment. As part of the baseline implementation, our aim is to devise an optimal navigation path through the environment illustrated in Figure \ref{fig:environment}, starting from the bottom center of the image and concluding at the top center of the image.

\section{Literature review}

Several research projects have focused on the development of autonomous robots capable of navigating diverse environments. However, there is a scarcity of research specifically addressing the navigation challenges faced by legged robots in indoor environments, particularly when relying on single-camera vision. Legged robots possess the ability to traverse uneven surfaces and overcome obstacles such as stairs, which are typically inaccessible to traditional wheeled robots.

While the common approach in the literature relies on LiDAR-based measurements for environment mapping and path planning, there has been limited research exploring vision-only methods for indoor navigation. Existing approaches often rely on preliminary techniques such as image contrast, which have a high probability of failure. In contrast, our paper proposes utilizing image segmentation with deep neural networks, which have shown significantly higher success rates in understanding the environment.

In the paper titled "Indoor Robot Navigation with Single Camera Vision" by Gini et al. \cite{gini2002indoor}, the authors explore indoor navigation using a wheeled robot equipped with a single camera. Their approach relies on image contrast to estimate ground and wall regions, employs a grid-based representation, and utilizes the A* search algorithm. Although this method demonstrates commendable progress, it exhibits limitations in accurately differentiating between multiple floor types and adequately perceiving obstacles, thus hindering its ability to assign varying costs for search.

Another notable work, "Development of an Autonomous Navigation System for an Indoor Service Robot Application" by Seo et al. \cite{seo2013development}, combines odometry and laser measurements to map the environment. Monte Carlo localization is employed for robot localization, and the A* algorithm is utilized for navigation planning. However, a drawback of this method is its incapability to maneuver around obstacles not detected by laser-based measurements, and the use of fixed costs for A* planning, which may not accurately reflect the optimal path in the environment.

\section{Dataset}

To develop and evaluate our navigation system, we built a dataset comprising images captured by the robot's RGB camera. The dataset encompasses varying home environments, including obstacles of different shapes arranged at different locations. The dataset used in this project includes 10 manually annotated environments, each containing various scenes and obstacles. This dataset is split into 2 logs and each log represents a set of diverse images with varying number of objects in a particular home environment. In addition, we have included 1200 sequential unlabeled images showcasing a moving robot in the given home environment. These images are specifically intended for testing purposes, allowing us to evaluate the navigation system's performance in dynamic scenarios. By incorporating sequential images, we aim to simulate real-world conditions and assess the system's ability to adapt and navigate effectively in changing environments.

\section{Methodology}

\subsection{Key steps}

\begin{figure}[H]
     \centering
     \begin{subfigure}[b]{0.3\textwidth}
         \centering
         \includegraphics[width=\textwidth, trim={0 0 0 5.5cm},clip]{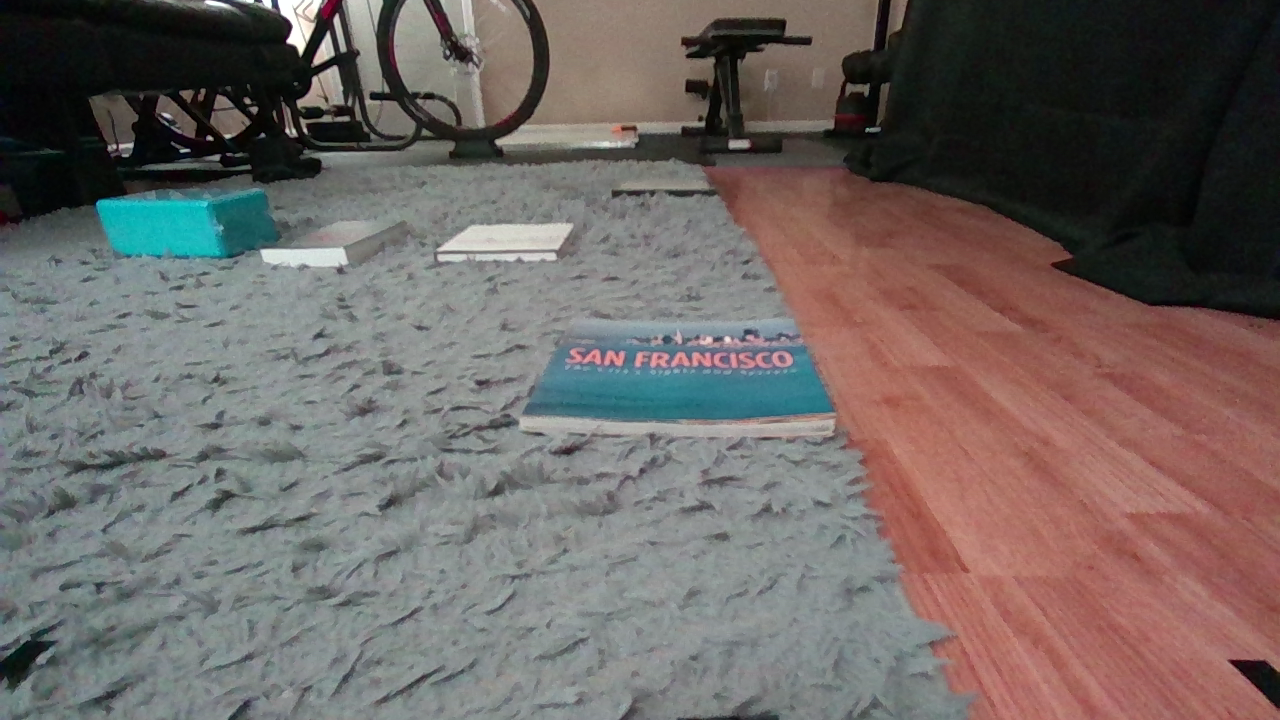}
         \caption{Camera frame from robot}
         \label{fig:frame}
     \end{subfigure}
     \hfill
     \begin{subfigure}[b]{0.3\textwidth}
         \centering
         \includegraphics[width=\textwidth]{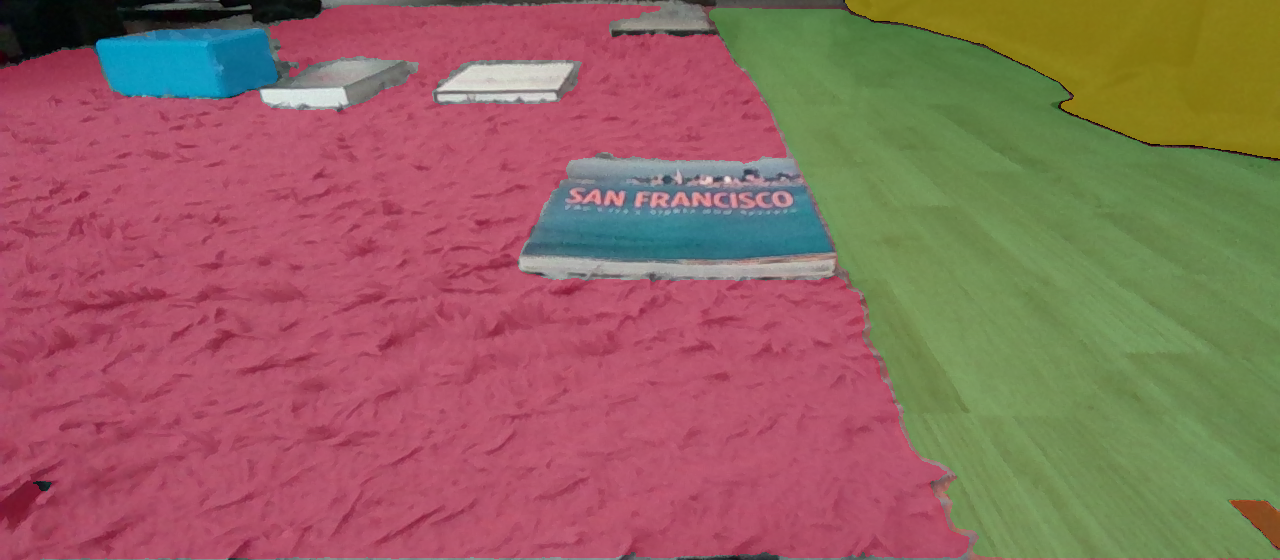}
         \caption{The segmented labels}
         \label{fig:segmented}
     \end{subfigure}
     \hfill
     \begin{subfigure}[b]{0.3\textwidth}
         \centering
         \includegraphics[width=\textwidth]{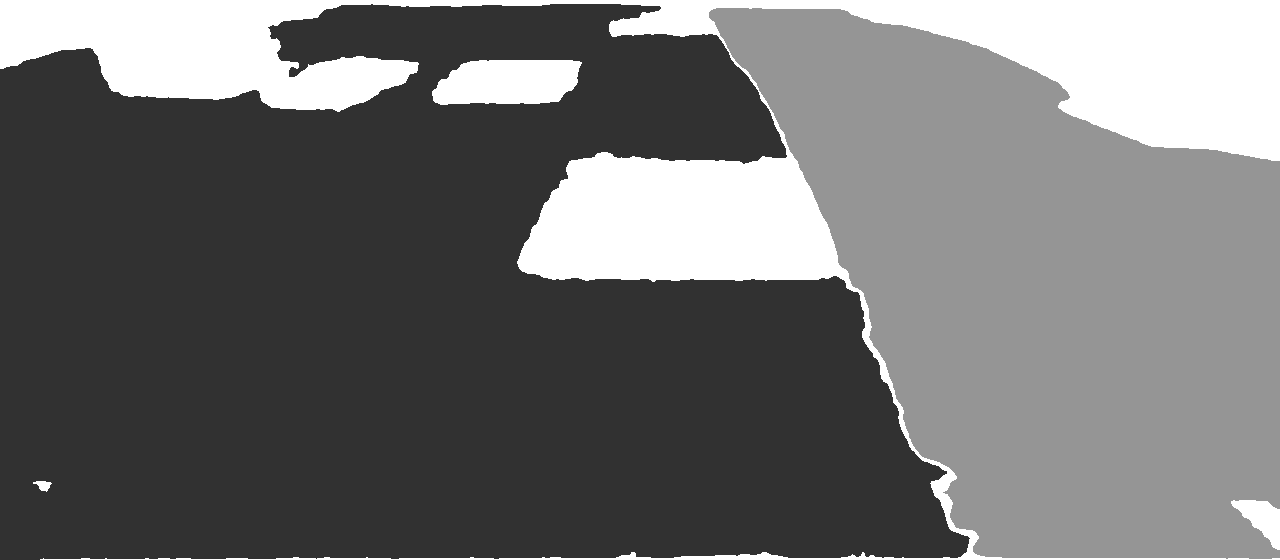}
         \caption{The cost map}
         \label{fig:cost}
     \end{subfigure}
        \caption{Finding cost}
        \label{fig:find_cost}
\end{figure}

The implementation involves several key steps for obstacle-avoidance navigation. First, we preprocess the color images by cropping them to focus on the relevant floor area (see Figure \ref{fig:frame}). Subsequently, we employ a state-of-the-art semantic segmentation network, known as Segment-Anything by Meta \cite{kirillov2023segment}, to obtain accurate floor and obstacle segmentation results (see Figure \ref{fig:segmented}). We assign costs to the segmented obstacles based on their characteristics and the robot's ability to traverse them (see Figure \ref{fig:cost}). The selection of costs plays a crucial role in determining the trajectory followed by the robot. As illustrated in Figure \ref{fig:cost}, lower costs are depicted by darker regions, while higher costs are represented by lighter regions. In the context of this project, we have made deliberate choices regarding cost assignment to various elements in the environment.

Specifically, we have assigned a relatively lower cost to walk on the carpeted surface, as the robot exhibits greater stability and maneuverability on this type of terrain. Conversely, a higher cost is allocated to walking on hardwood floors due to the tendency of the 3D-printed foot of the robot to experience slippage in such conditions. Moreover, objects such as books and other obstacles within the environment are treated as impediments and are assigned a significantly higher cost. This strategic cost assignment effectively encourages the robot to circumvent these obstacles during path planning, promoting efficient navigation through the environment.

\begin{figure}[H]
     \centering
     \begin{subfigure}[b]{0.28\textwidth}
         \centering
         \includegraphics[width=\textwidth]{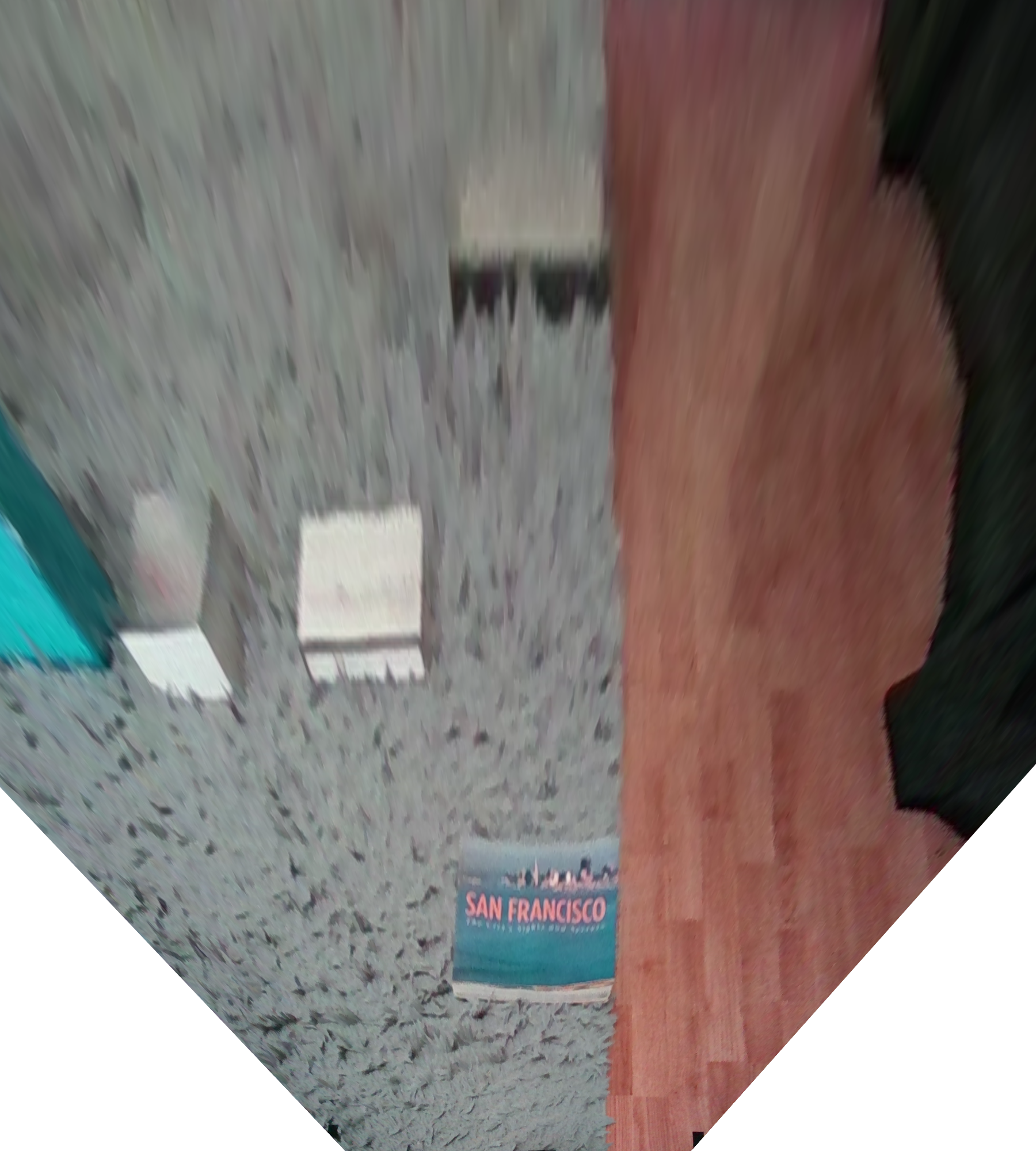}
         \caption{Camera frame in BEV view}
         \label{fig:color_bev}
     \end{subfigure}
     \hfill
     \begin{subfigure}[b]{0.28\textwidth}
         \centering
         \includegraphics[width=\textwidth]{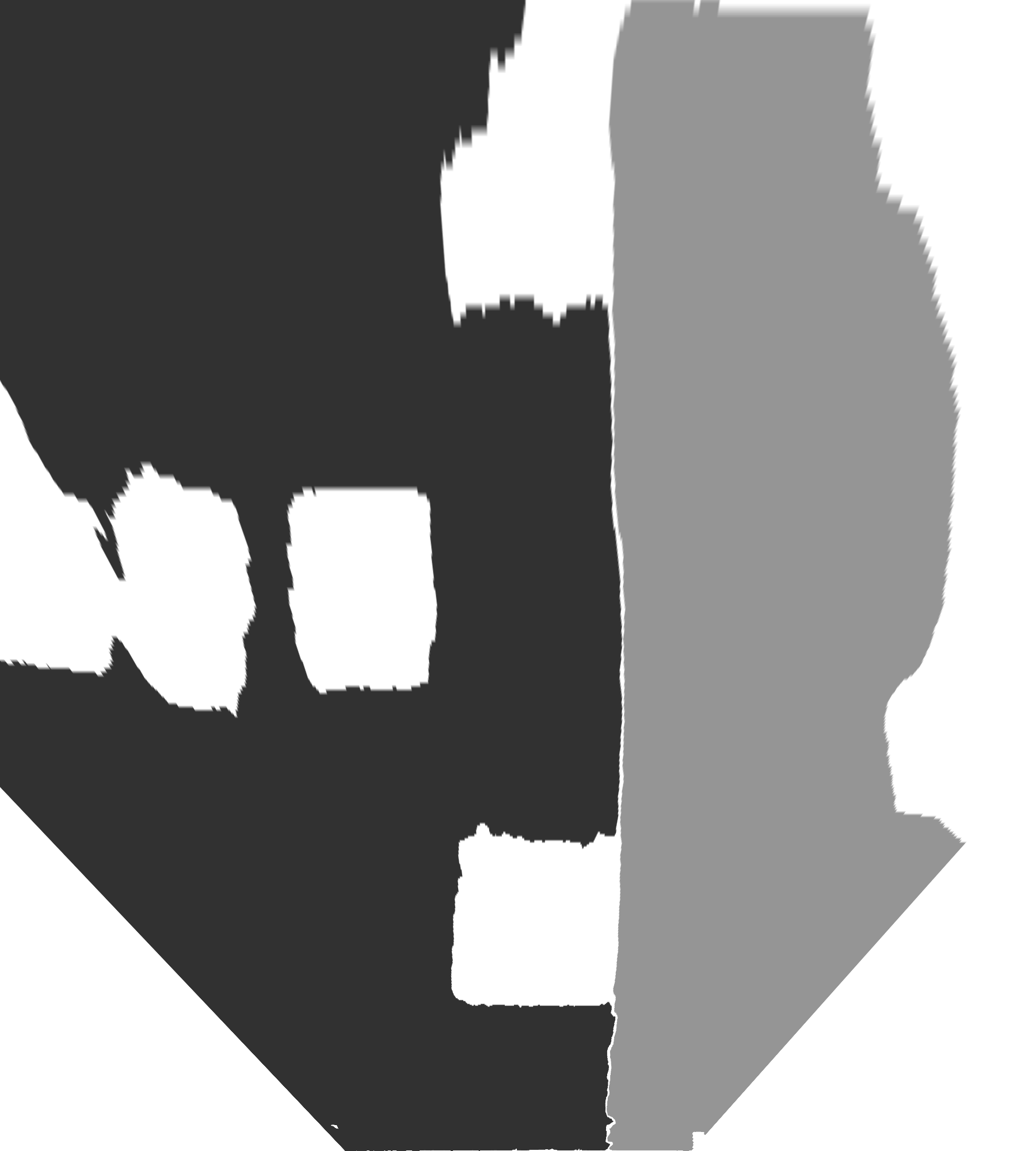}
         \caption{The cost map in BEV view}
         \label{fig:cost_bev}
     \end{subfigure}
     \hfill
     \begin{subfigure}[b]{0.29\textwidth}
         \centering
         \includegraphics[width=\textwidth]{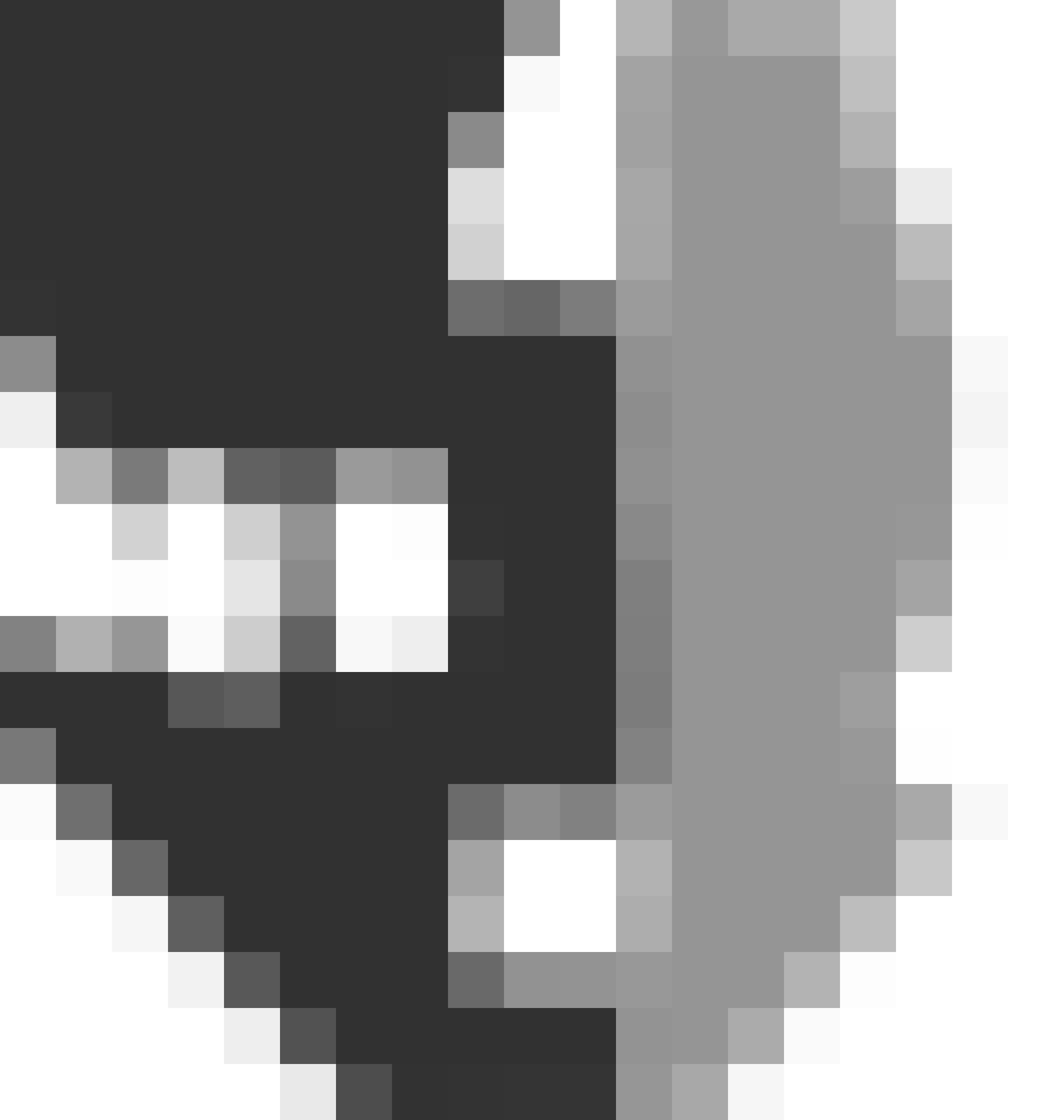}
         \caption{The cost grid in BEV}
         \label{fig:cost_grid}
     \end{subfigure}
        \caption{Cost grid from the cost map}
        \label{fig:cost_grid_steps}
\end{figure}

The obtained cost image is subsequently transformed into a birds-eye view (BEV) perspective, as depicted in Figure \ref{fig:cost_bev}, alongside the corresponding color image shown in Figure \ref{fig:color_bev}. This transformation process involves calculating the mapping between the perspective view and the BEV space. To achieve this, four points in the perspective image are identified and matched to their corresponding positions in the BEV space. By employing homography, a transformation matrix is then computed to map the two planes. To enable efficient search operations, a pixel-to-millimeter ratio of 1:1 is employed, wherein one pixel in the BEV space corresponds to one millimeter on the ground. Leveraging this transformation matrix, the perspective image is remapped to generate the BEV image, as exemplified in the transition from Figure \ref{fig:frame} to Figure \ref{fig:color_bev}.

The resulting BEV cost map, presented in Figure \ref{fig:cost_bev}, is then utilized to construct a 19x20 cost grid, as illustrated in Figure \ref{fig:cost_grid}. This involves calculating the mean cost value within each grid cell, which corresponds to a 100x100mm area on the ground. The choice of this cell size is influenced by the size of the cost grid as well as the width of the robot. Subsequently, the A* algorithm is employed to determine the optimal path with the lowest cost, taking into account both obstacle avoidance and efficient navigation toward the destination point. 

\subsection{A* setup}

In the grid-based A* algorithm, the state is represented by the current location, while the available actions correspond to movement in the four cardinal directions. Initially, a cost map is generated, assigning very high costs to all cells. As the algorithm explores the states, these costs are gradually updated. Successor states are determined based on the current state and selected actions, and they are added to a priority queue sorted by their associated costs. To introduce heuristics into the cost calculations for A*, we have incorporated the Manhattan distance between the current state and the destination as the heuristic measure. This choice of heuristic is motivated by its consistency, as it typically underestimates the actual cost required to reach the destination. The algorithm proceeds by exploring the state with the lowest cost in the queue, updating the priority queue and cost map based on the determined successors, and repeating this process until the destination state is reached. 

\subsection{Real-time optimization}

The performance bottleneck of the method lies in the semantic segmentation step, which currently takes over 15 seconds to process each frame. To overcome this limitation and achieve real-time optimization, three modifications were implemented:

Firstly, the model was quantized from FP32 to INT8 precision, resulting in a negligible loss of segmentation performance. Secondly, the input image resolution was reduced from 1280x720 to 640x360, impacting segmentation accuracy at long-range but maintaining effectiveness at shorter distances. Lastly, the vit\_b model with the smallest backbone in the Segment Anything models was utilized, causing only a minor reduction in segmentation accuracy.

These optimizations resulted in the model being able to process each image in well under a second on the Nvidia Jetson Xavier NX onboard the robot, significantly improving real-time performance.

\section{Evaluation Metric}

To evaluate the effectiveness of our obstacle-avoidance navigation system, we employ both qualitative and quantitative metrics. Qualitatively, we visually assess the generated navigation path overlaid on the BEV color image to verify its adherence to the desired trajectory and successful obstacle avoidance. Quantitatively, we compare the generated paths with the manually-annotated data to benchmark the system's performance.

To further evaluate the effectiveness of our obstacle-avoidance navigation system, we will also qualitatively evaluate the system's performance on a dataset of 1200 sequential images showing a moving robot in the environment. By visually assessing the inference results on these images, we can gain insights into the system's ability to effectively navigate and avoid obstacles in a dynamic environment. Additionally, this qualitative evaluation will provide valuable feedback on the system's performance in real-world scenarios, complementing the quantitative metrics previously mentioned.

\section{Results and Analysis}

At the present stage, we have observed excellent results with the A* navigation system. 

\begin{figure}[H]
     \centering
     \begin{subfigure}[b]{0.22\textwidth}
         \centering
         \includegraphics[width=\textwidth]{images/frame_cost_grid.png}
         \caption{The BEV cost grid}
         \label{fig:cost_grid_2}
     \end{subfigure}
     \hfill
     \begin{subfigure}[b]{0.24\textwidth}
         \centering
         \includegraphics[width=\textwidth]{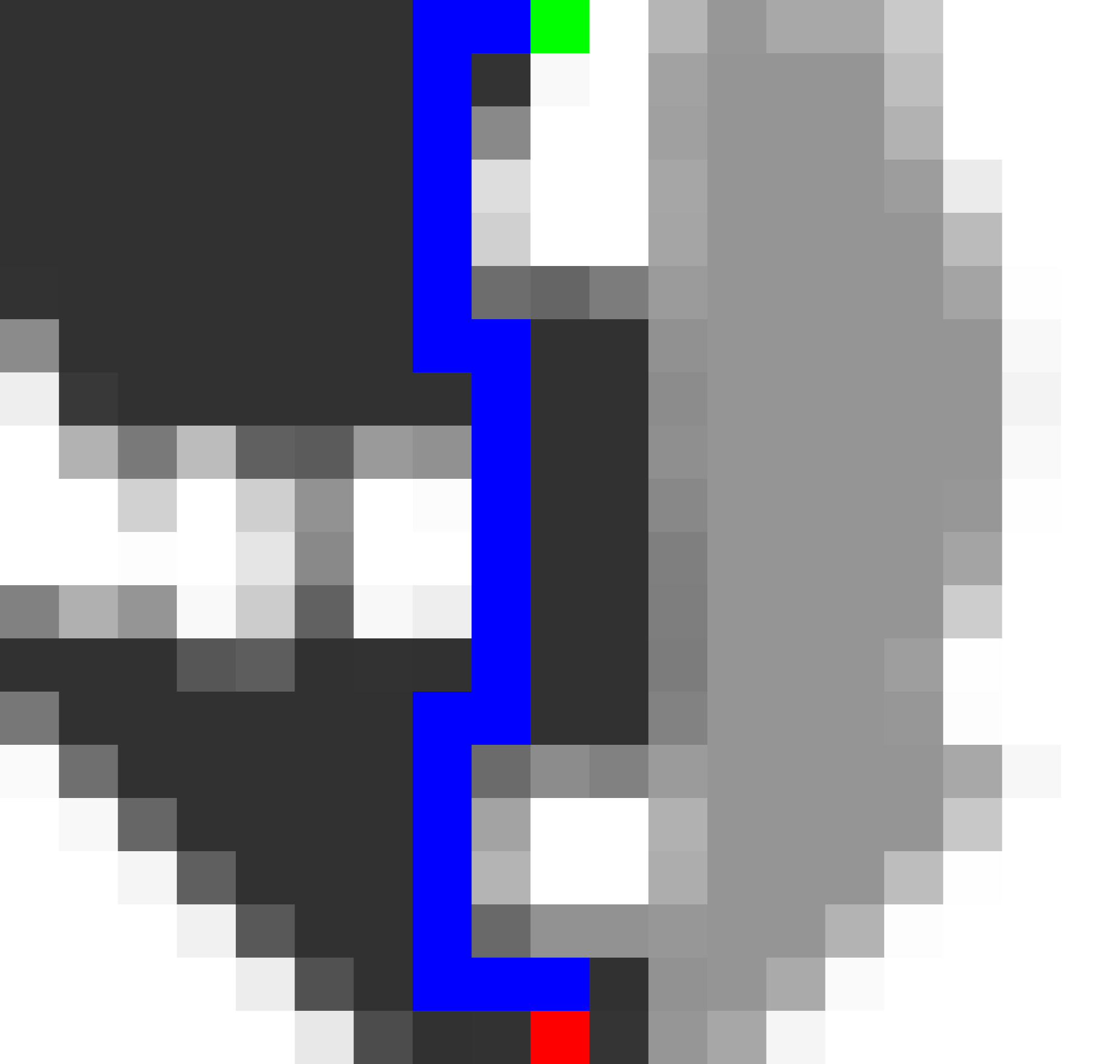}
         \caption{The cost grid with A*}
         \label{fig:cost_grid_a_star}
     \end{subfigure}
     \hfill
     \begin{subfigure}[b]{0.4\textwidth}
         \centering
         \includegraphics[width=\textwidth]{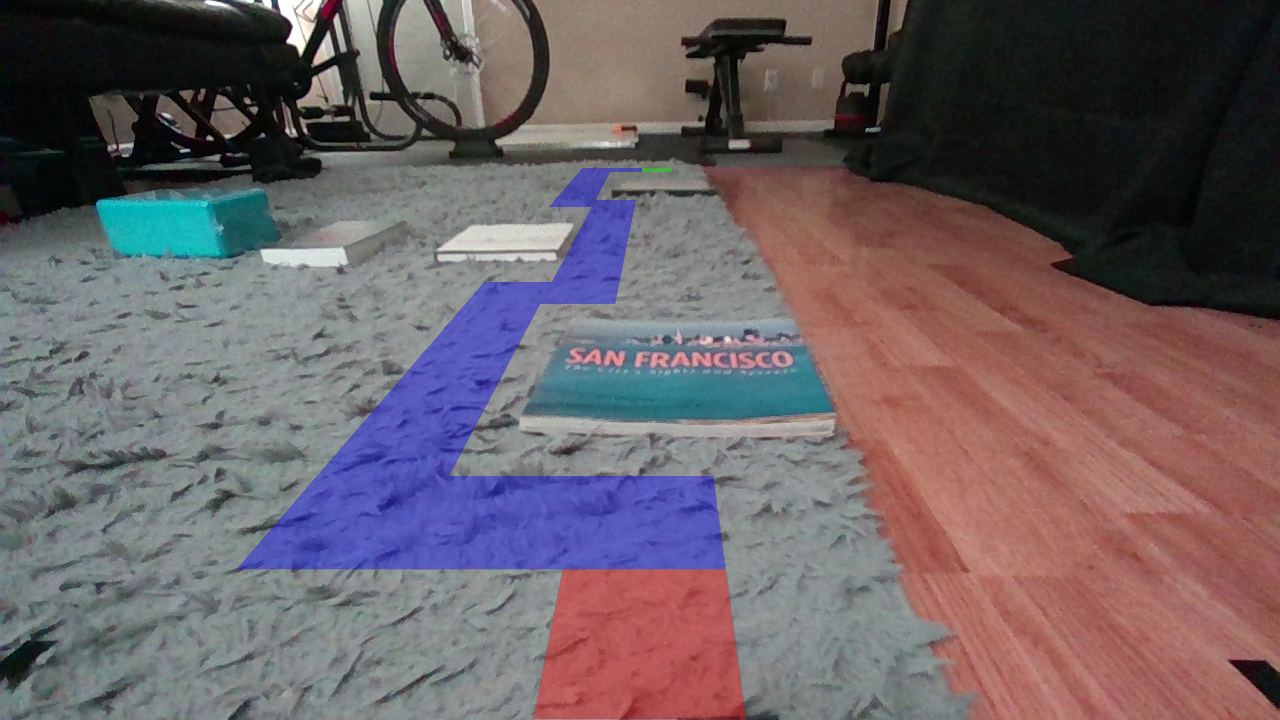}
         \caption{A* path in BEV}
         \label{fig:bev_a_star}
     \end{subfigure}
        \caption{Results}
        \label{fig:a_star_results}
\end{figure}

Figure \ref{fig:a_star_results} above depicts the outcomes obtained from the A* search implementation. Figures \ref{fig:cost_grid_2} and \ref{fig:cost_grid_a_star} illustrate the cost grid and the corresponding optimal low-cost path generated using the A* algorithm. It is important to note that the origin of navigation in the image is located at the bottom center (indicated by the red marker in \ref{fig:cost_grid_a_star}), while the destination is positioned at the top center (indicated by the green marker in \ref{fig:cost_grid_a_star}). 

Moreover, by computing the inverse transformation matrix, we can convert the bird's-eye view (BEV) to the perspective view. This enables us to transform the results into the perspective view and overlay them onto the original image for visualization. Figure \ref{fig:a_star_results} demonstrates this overlay, showcasing the path computed based on vision-based sensors alone. The visual representation highlights the system's ability to perform accurate obstacle detection and avoidance through qualitative analysis.

\subsection{Qualitative Evaluation}

\begin{figure}[H]
\centering
    \begin{tabular}{cccc}
        \subfloat[No Obstacle Image]{\includegraphics[width = 1.2in]{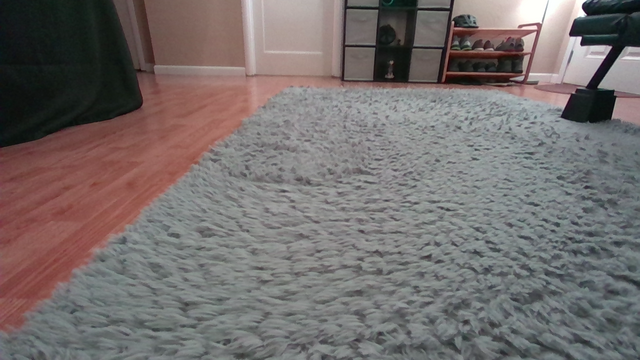}} &
        \subfloat[Cost Map]{\includegraphics[width = 1.2in]{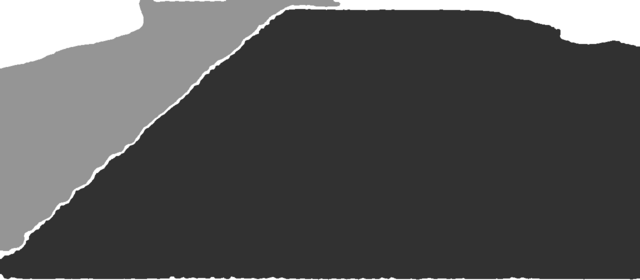}} &
        \subfloat[Cost Grid Search]{\includegraphics[width = 1.2in]{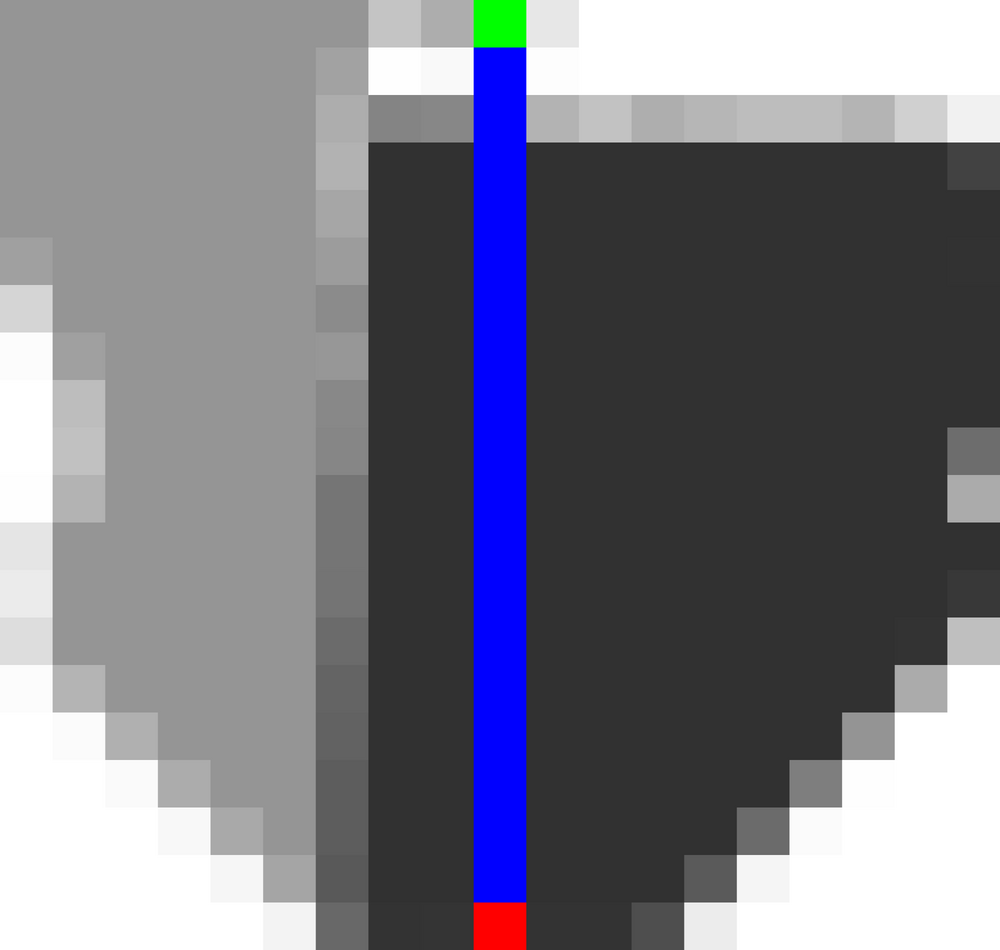}} &
        \subfloat[Search Path]{\includegraphics[width = 1.2in]{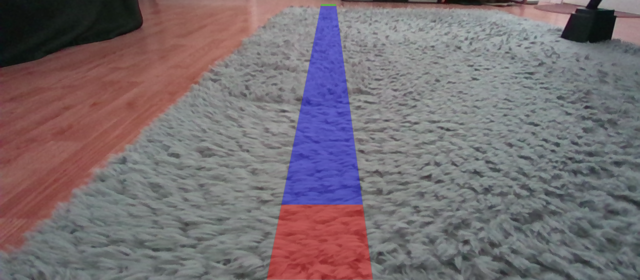}} \\
        \subfloat[1 Obstacle Image]{\includegraphics[width = 1.2in]{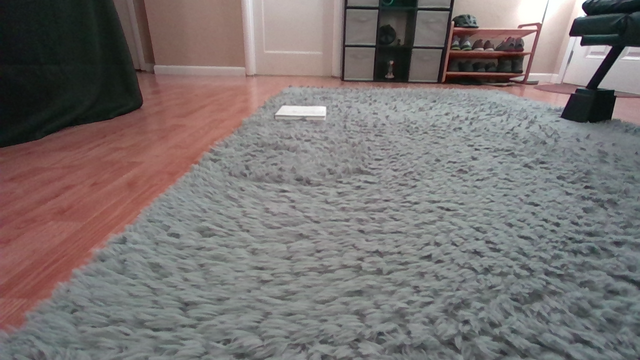}} &
        \subfloat[Cost Map]{\includegraphics[width = 1.2in]{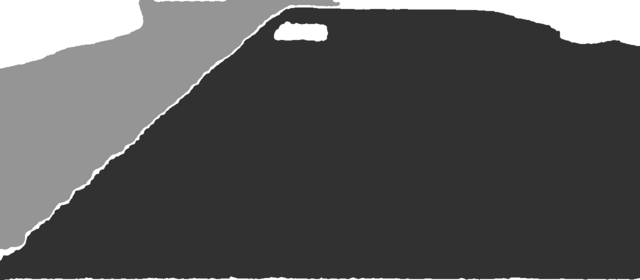}} &
        \subfloat[Cost Grid Search]{\includegraphics[width = 1.2in]{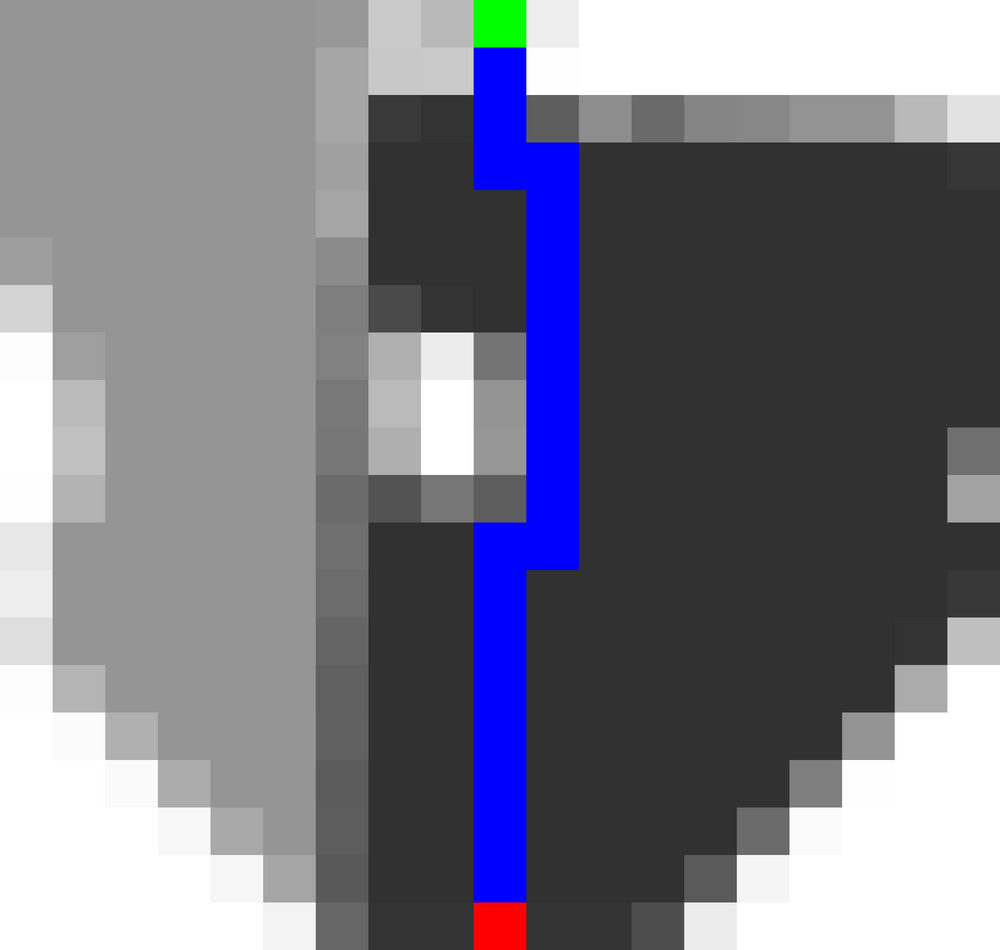}} &
        \subfloat[Search Path]{\includegraphics[width = 1.2in]{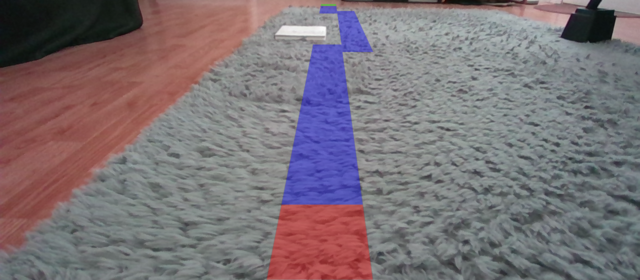}} \\
        \subfloat[3 Obstacle Image]{\includegraphics[width = 1.2in]{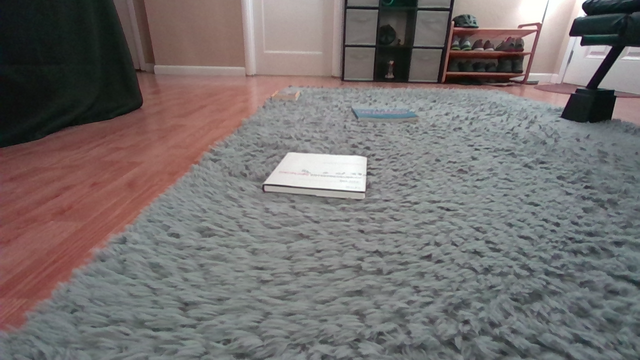}} &
        \subfloat[Cost Map]{\includegraphics[width = 1.2in]{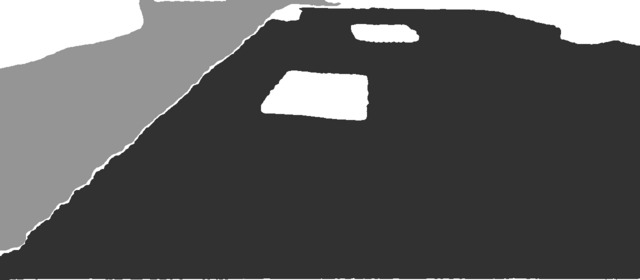}} &
        \subfloat[Cost Grid Search]{\includegraphics[width = 1.2in]{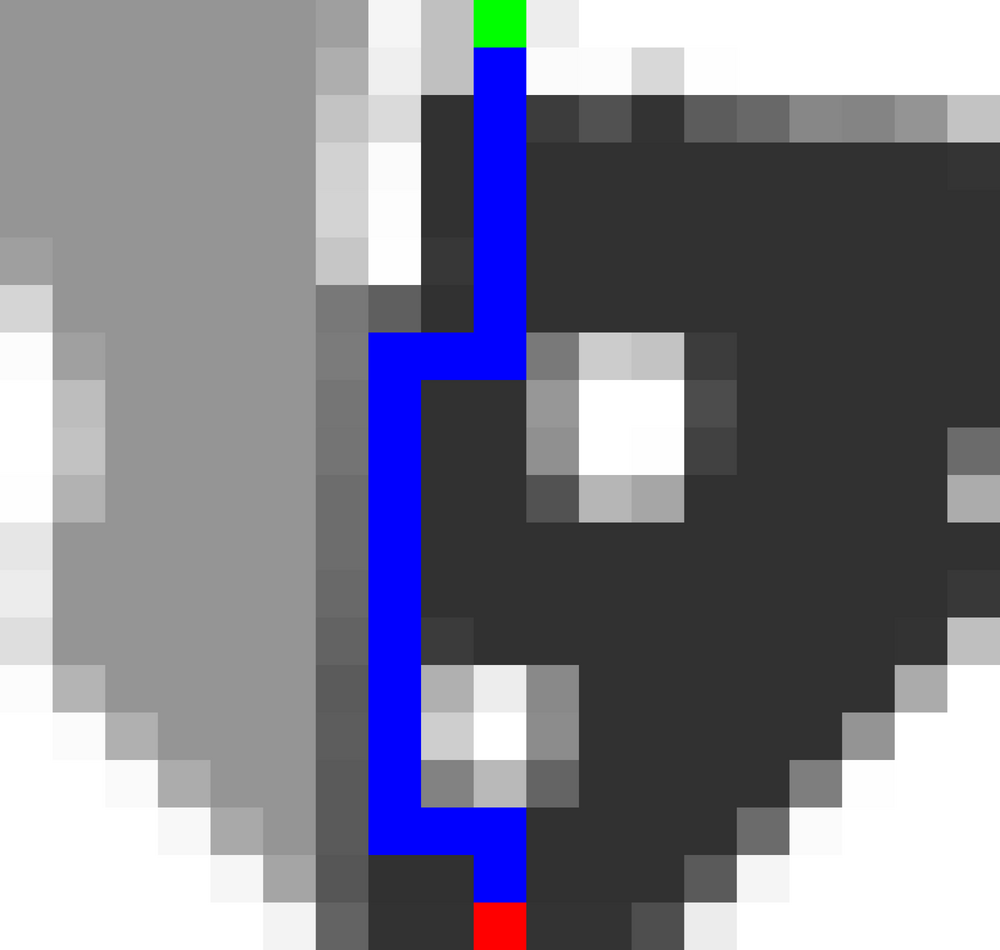}} &
        \subfloat[Search Path]{\includegraphics[width = 1.2in]{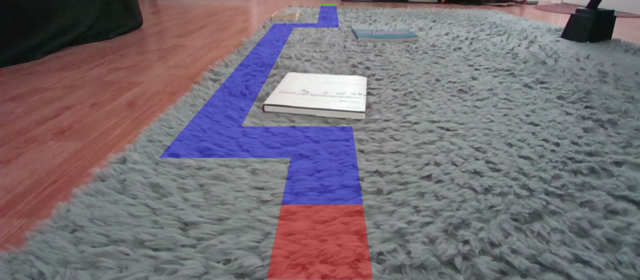}} \\
        \subfloat[5 Obstacle Image]{\includegraphics[width = 1.2in]{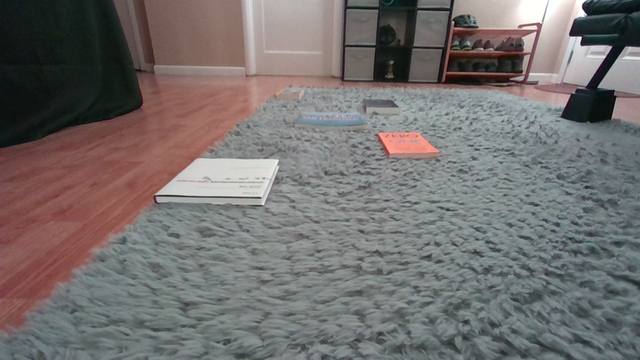}} &
        \subfloat[Cost Map]{\includegraphics[width = 1.2in]{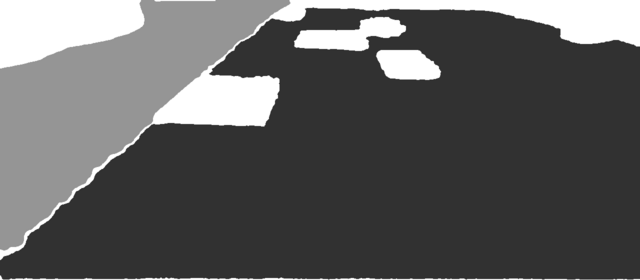}} &
        \subfloat[Cost Grid Search]{\includegraphics[width = 1.2in]{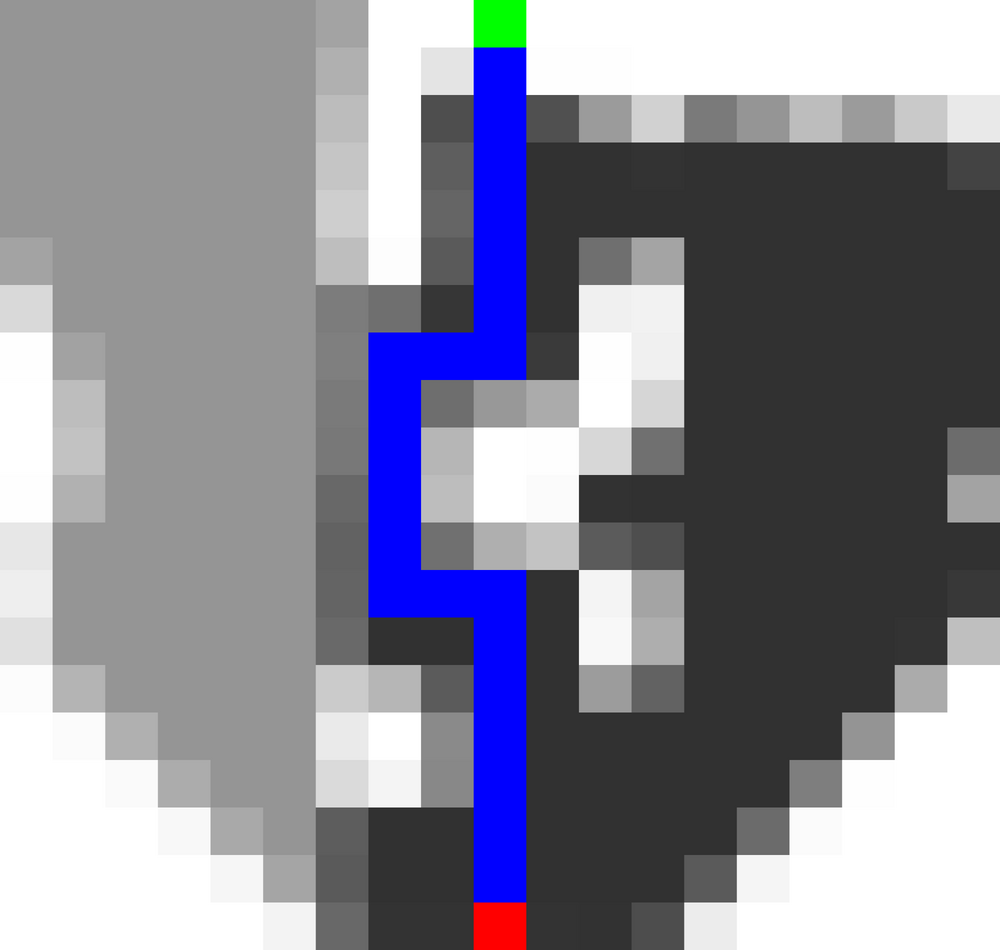}} &
        \subfloat[Search Path]{\includegraphics[width = 1.2in]{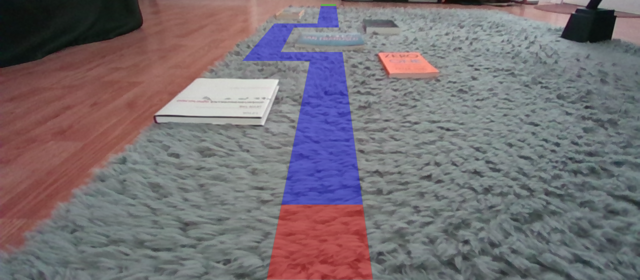}} \\
        \subfloat[Blocked Path Image]{\includegraphics[width = 1.2in]{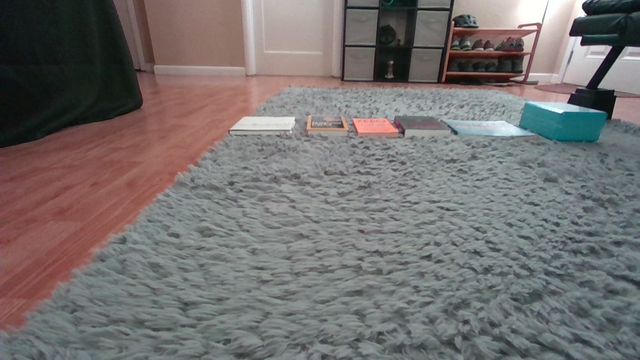}} &
        \subfloat[Cost Map]{\includegraphics[width = 1.2in]{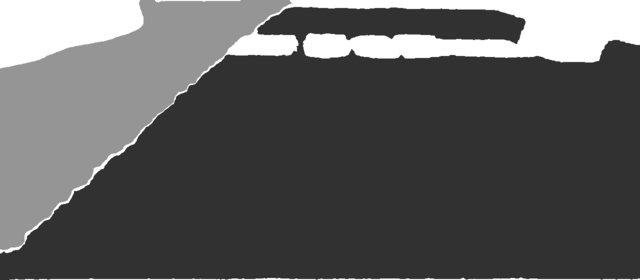}} &
        \subfloat[Cost Grid Search]{\includegraphics[width = 1.2in]{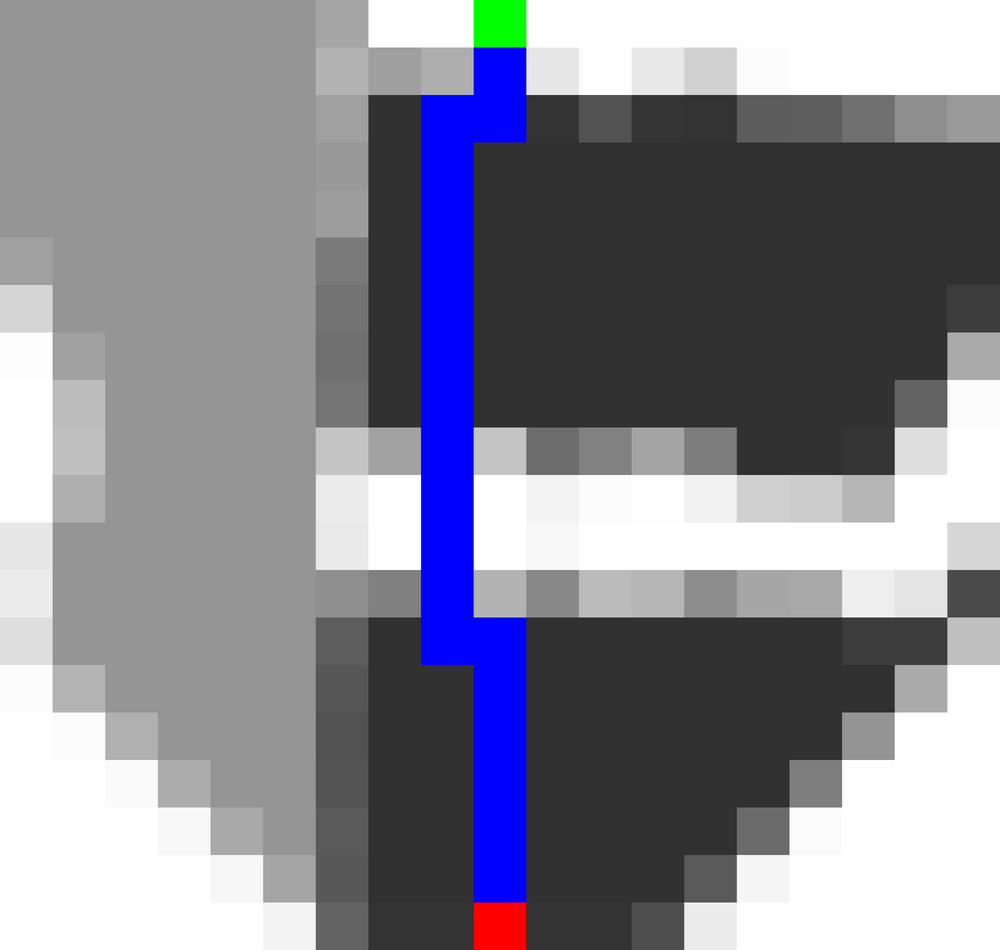}\label{fig:blocked_1}} &
        \subfloat[Search Path]{\includegraphics[width = 1.2in]{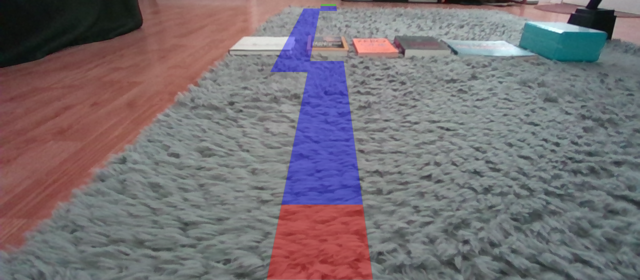}} 
    \end{tabular}
    \caption{Results with Log 1 of the Dataset}
    \label{fig:log_1_results}
\end{figure}

\begin{figure}[H]
\centering
    \begin{tabular}{cccc}
        \subfloat[No Obstacle Image]{\includegraphics[width = 1.2in]{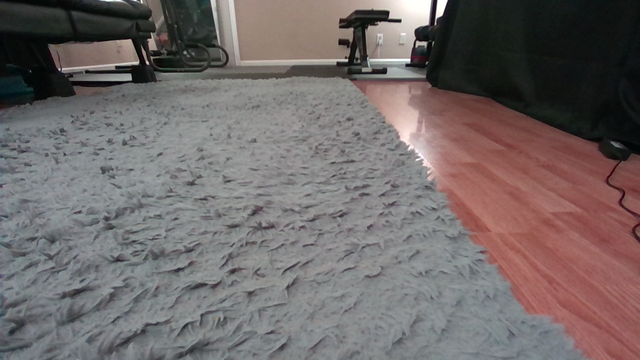}} &
        \subfloat[Cost Map]{\includegraphics[width = 1.2in]{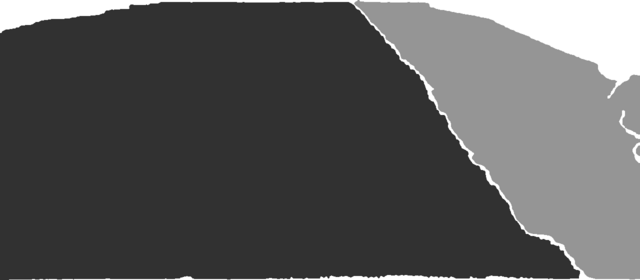}} &
        \subfloat[Cost Grid Search]{\includegraphics[width = 1.2in]{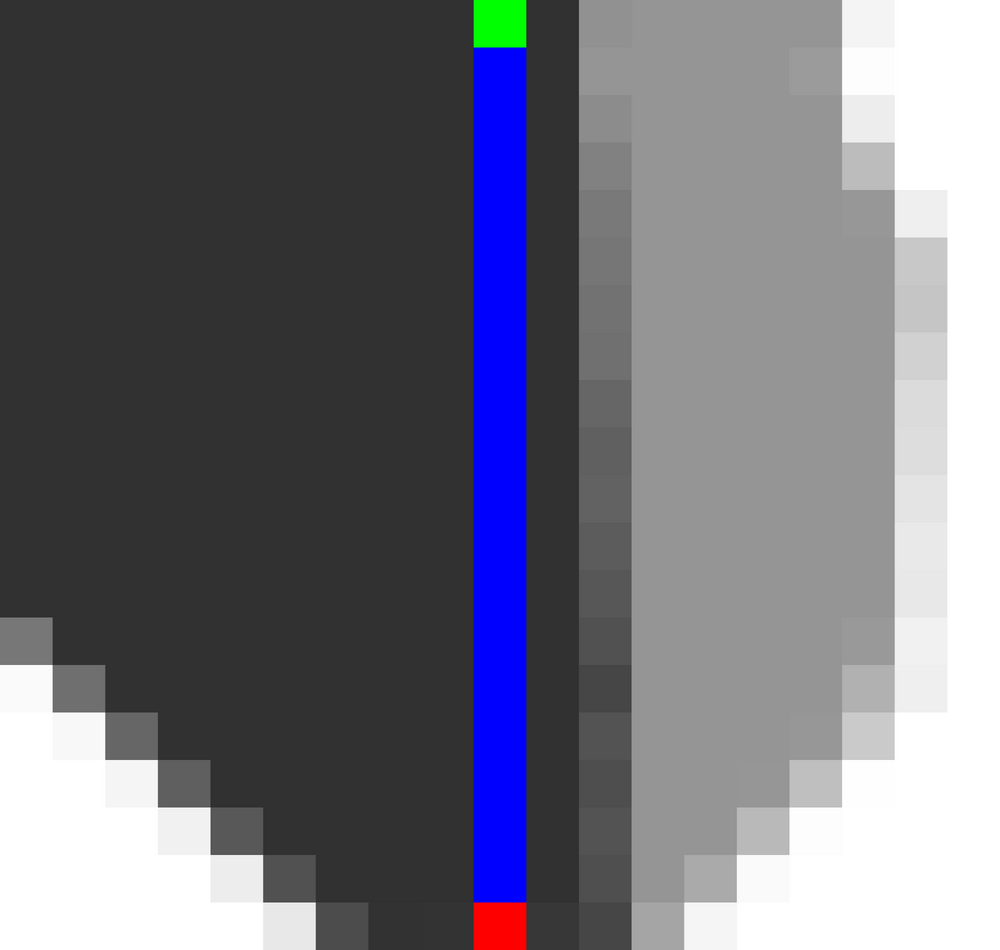}} &
        \subfloat[Search Path]{\includegraphics[width = 1.2in]{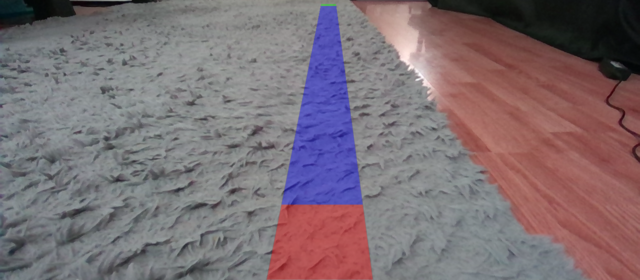}} \\
        \subfloat[1 Obstacle Image]{\includegraphics[width = 1.2in]{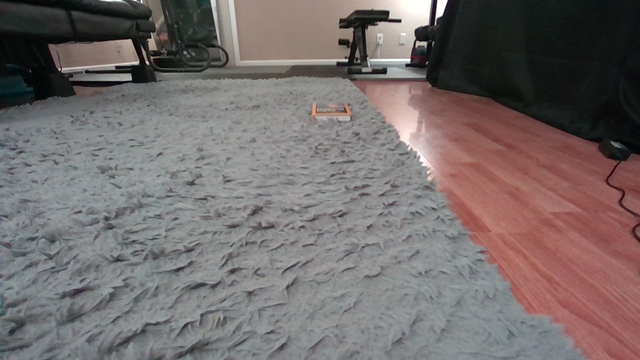}} &
        \subfloat[Cost Map]{\includegraphics[width = 1.2in]{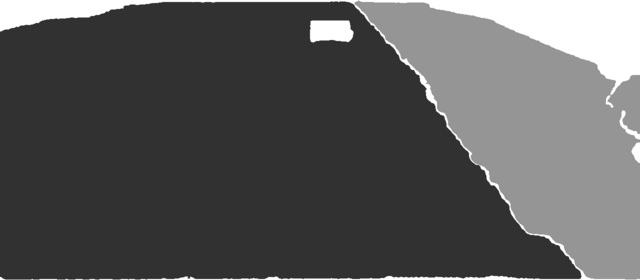}} &
        \subfloat[Cost Grid Search]{\includegraphics[width = 1.2in]{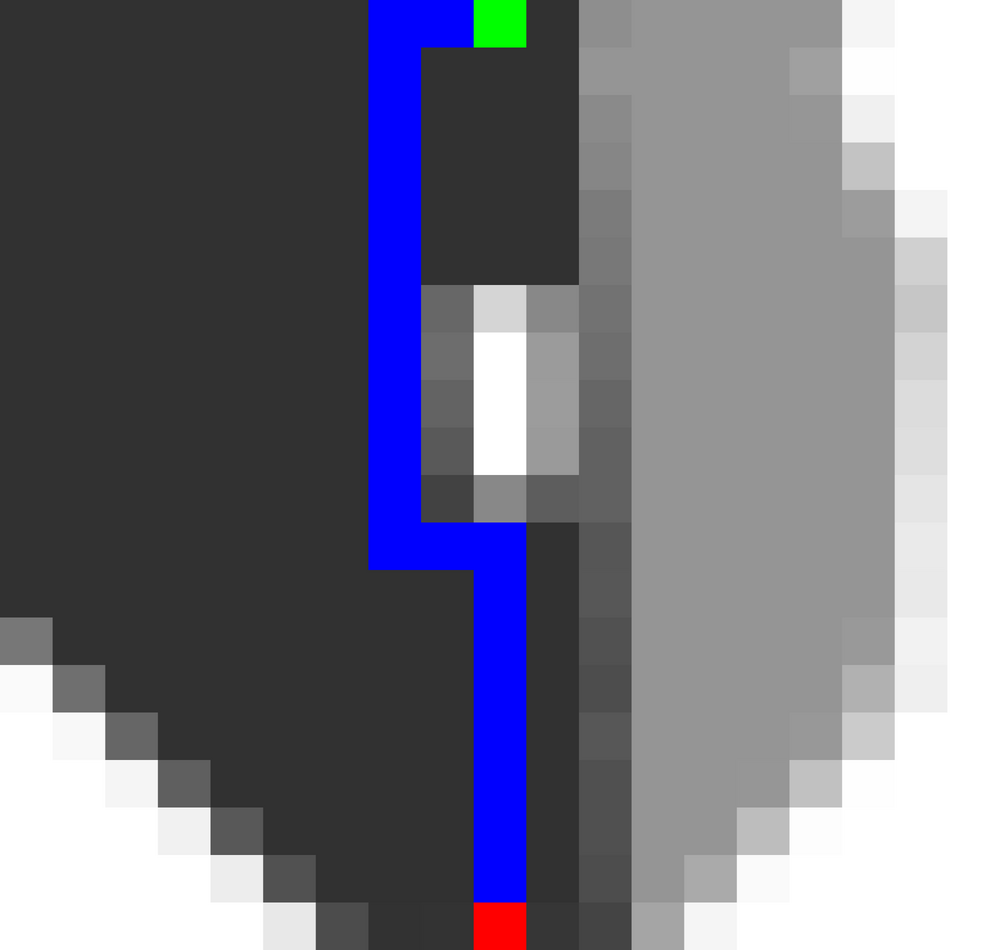}} &
        \subfloat[Search Path]{\includegraphics[width = 1.2in]{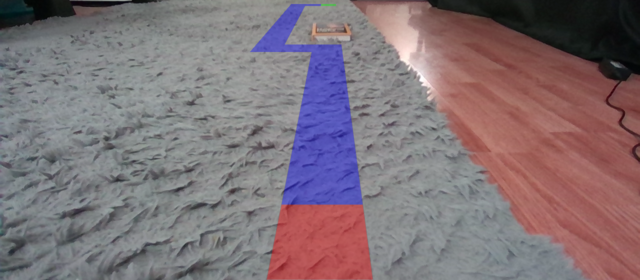}} \\
        \subfloat[3 Obstacle Image]{\includegraphics[width = 1.2in]{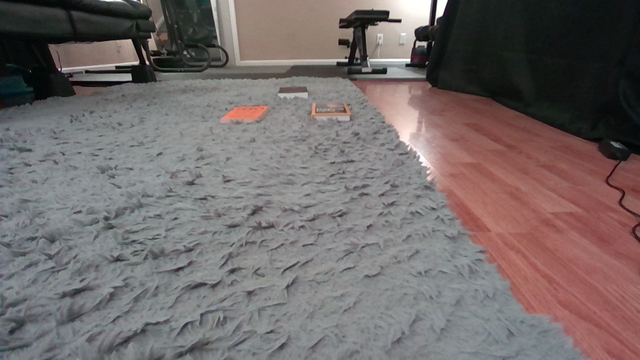}} &
        \subfloat[Cost Map]{\includegraphics[width = 1.2in]{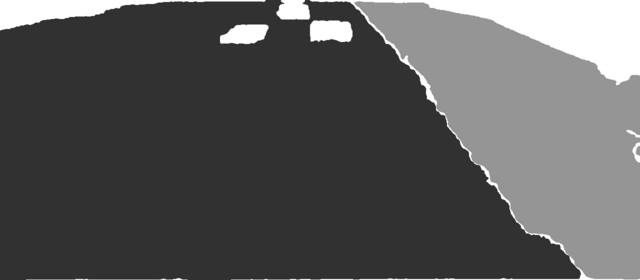}} &
        \subfloat[Cost Grid Search]{\includegraphics[width = 1.2in]{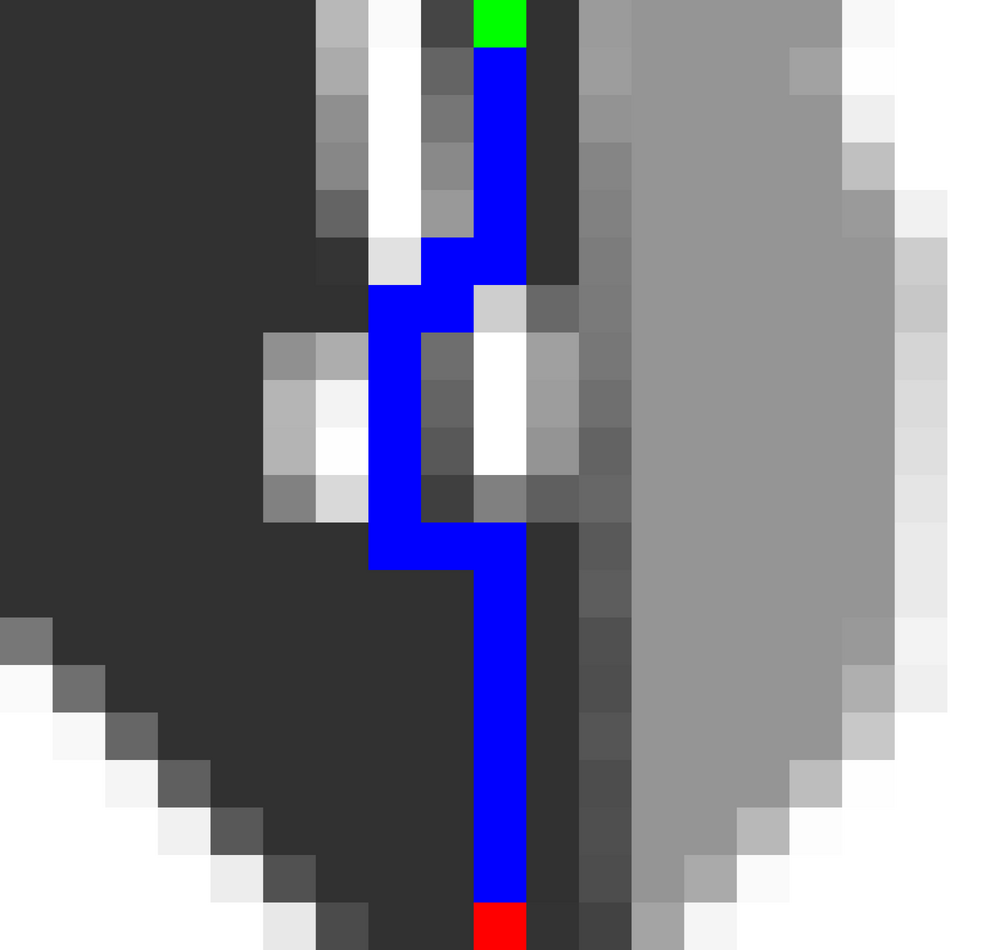}} &
        \subfloat[Search Path]{\includegraphics[width = 1.2in]{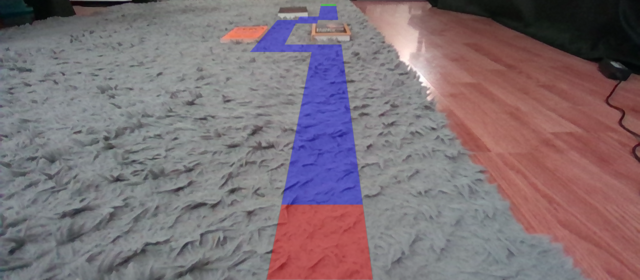}} \\
        \subfloat[5 Obstacle Image]{\includegraphics[width = 1.2in]{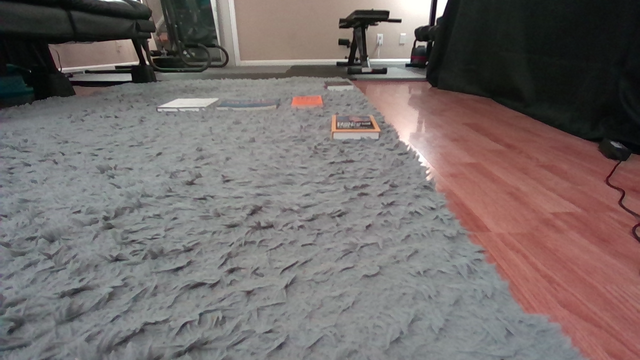}} &
        \subfloat[Cost Map]{\includegraphics[width = 1.2in]{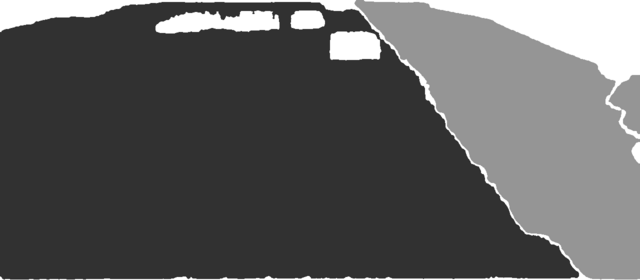}} &
        \subfloat[Cost Grid Search]{\includegraphics[width = 1.2in]{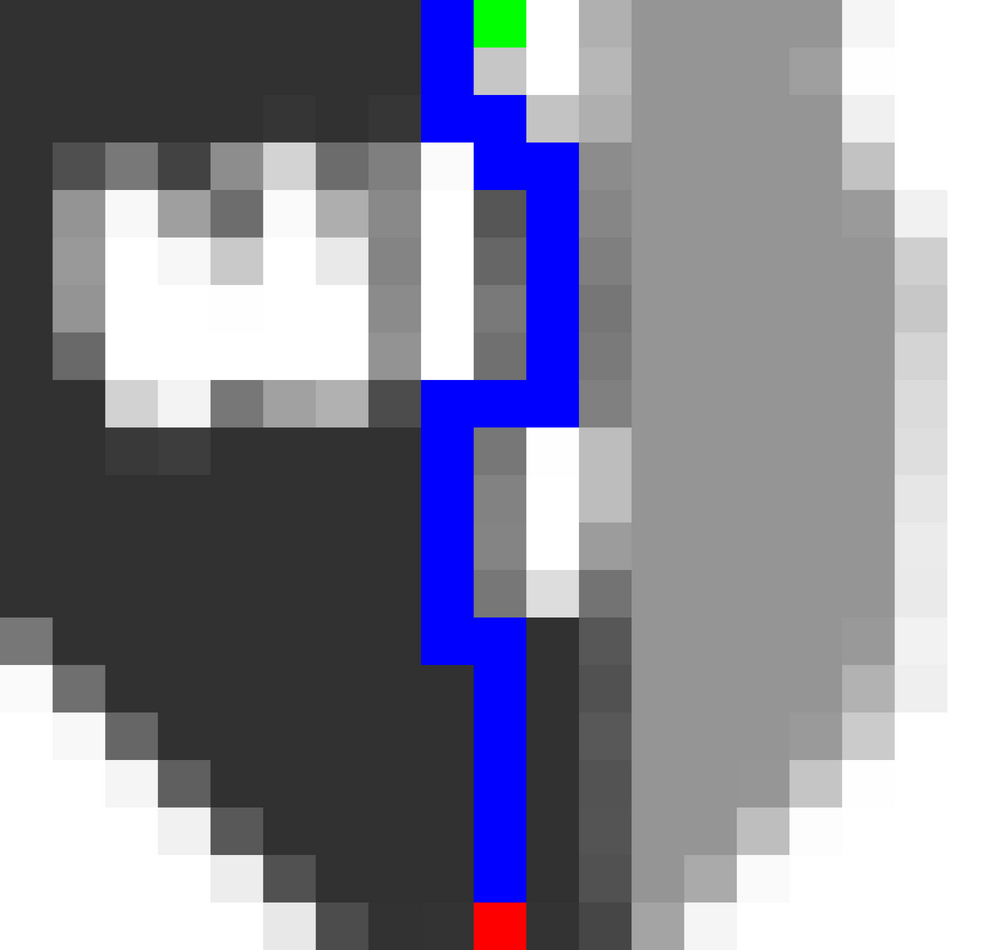}\label{fig:5_obst_2}} &
        \subfloat[Search Path]{\includegraphics[width = 1.2in]{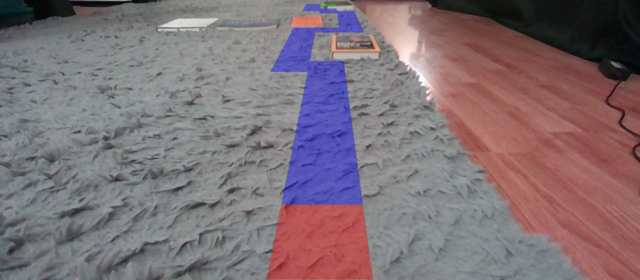}} \\
        \subfloat[Blocked Path Image]{\includegraphics[width = 1.2in]{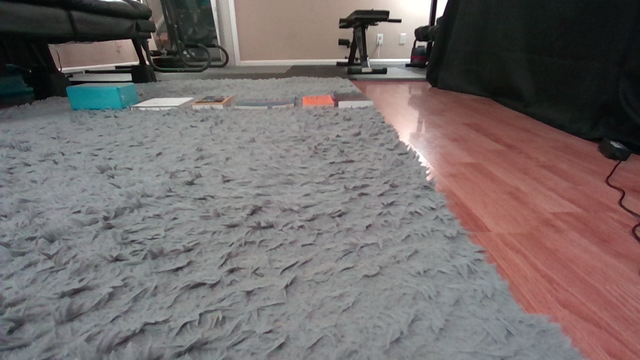}} &
        \subfloat[Cost Map]{\includegraphics[width = 1.2in]{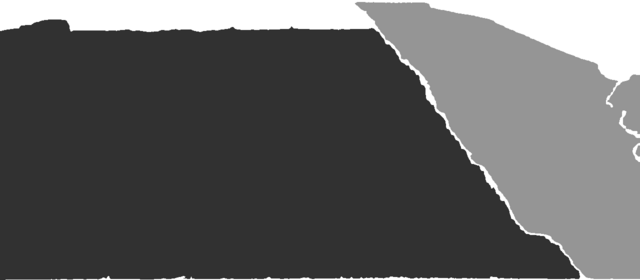}} &
        \subfloat[Cost Grid Search]{\includegraphics[width = 1.2in]{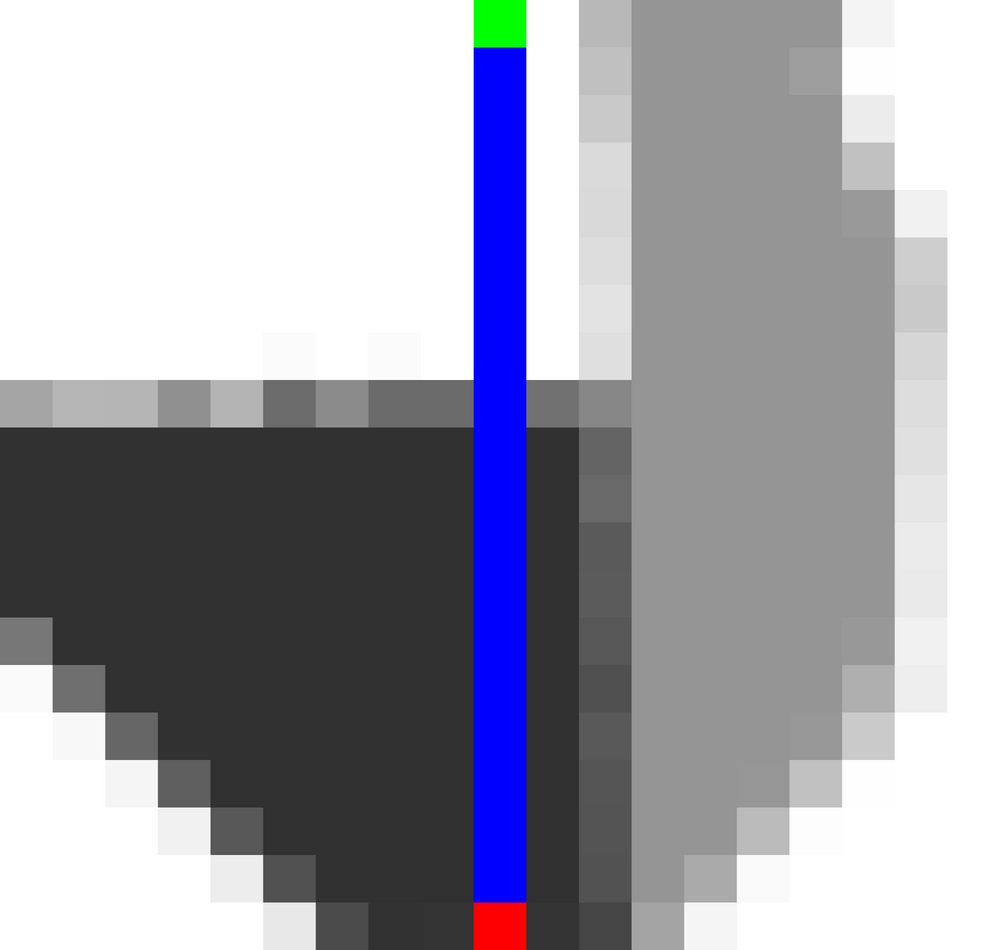}\label{fig:blocked_2}} &
        \subfloat[Search Path]{\includegraphics[width = 1.2in]{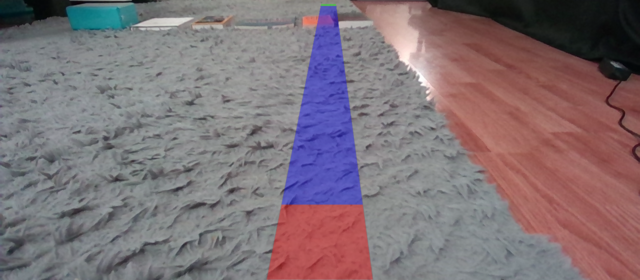}} 
    \end{tabular}
    \caption{Results with Log 2 of the Dataset}
    \label{fig:log_2_results}
\end{figure}

Qualitative evaluation involves visually analyzing the navigation paths generated by the obstacle-avoidance system overlaid on environment images. This helps identify strengths while also highlighting any unexpected behaviors during navigation. Figures \ref{fig:log_1_results} and \ref{fig:log_2_results} below represent the dataset used for evaluation, along with their corresponding qualitative assessments. 

\newpage

\subsection{Quantitative Evaluation}

We performed manual labeling for each step in a dataset comprising 10 images, encompassing the entire 19x20 grid. These labels were then compared against the paths generated by the search algorithm. The results of this comparison are presented in the following tables, with each table representing a specific log.

\begin{table}[H]
\begin{tabular}{|l||l|l|l|l|}
\hline
Log 1 scene        & Steps in result & Steps in label & Matching steps & Different steps \\
\hline
\hline
No objects   & 18              & 18             & 18    & 0                \\
\hline
1 object     & 20              & 20             & 20    & 0                \\
\hline
3 objects    & 22              & 22             & 22     & 0               \\
\hline
5 objects    & 22              & 22             & 22     & 0               \\
\hline
Blocked path & 20              & 26             & 16     & 10              \\
\hline
\end{tabular}
\vspace{2mm}
\caption{Results with Log 1 of the Dataset}
\end{table}

\begin{table}[H]
\begin{tabular}{|l||l|l|l|l|}
\hline
Log 2 scene        & Steps in result & Steps in label & Matching steps & Different steps \\
\hline
\hline
No objects   & 18              & 18             & 18    & 0                \\
\hline
1 object     & 22              & 22             & 22    & 0                \\
\hline
3 objects    & 22              & 22             & 22     & 0               \\
\hline
5 objects    & 26              & 26             & 25     & 1               \\
\hline
Blocked path & 18              & 24             & 14     & 10              \\
\hline
\end{tabular}
\vspace{2mm}
\caption{Results with Log 2 of the Dataset}
\end{table}

\subsection{Qualitative Evaluation of Test Data}

\begin{figure}[H]
     \centering
     \begin{subfigure}[b]{0.3\textwidth}
         \centering
         \includegraphics[width=\textwidth]{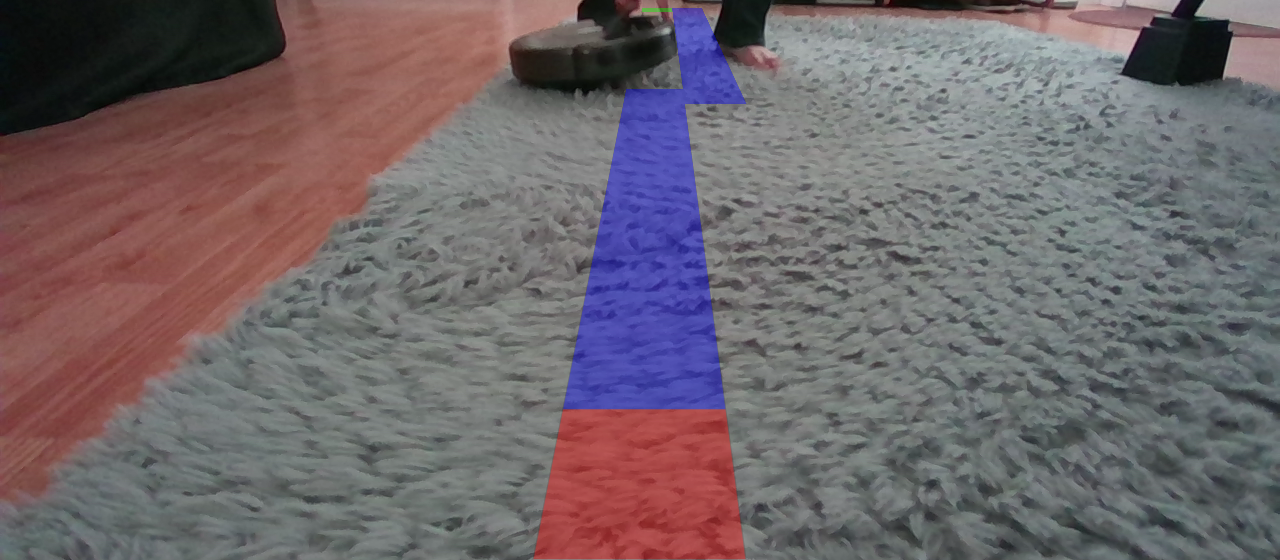}
         \caption{Find Path Between Obstacles}
         \label{fig:path_between}
     \end{subfigure}
     \hfill
     \begin{subfigure}[b]{0.3\textwidth}
         \centering
         \includegraphics[width=\textwidth]{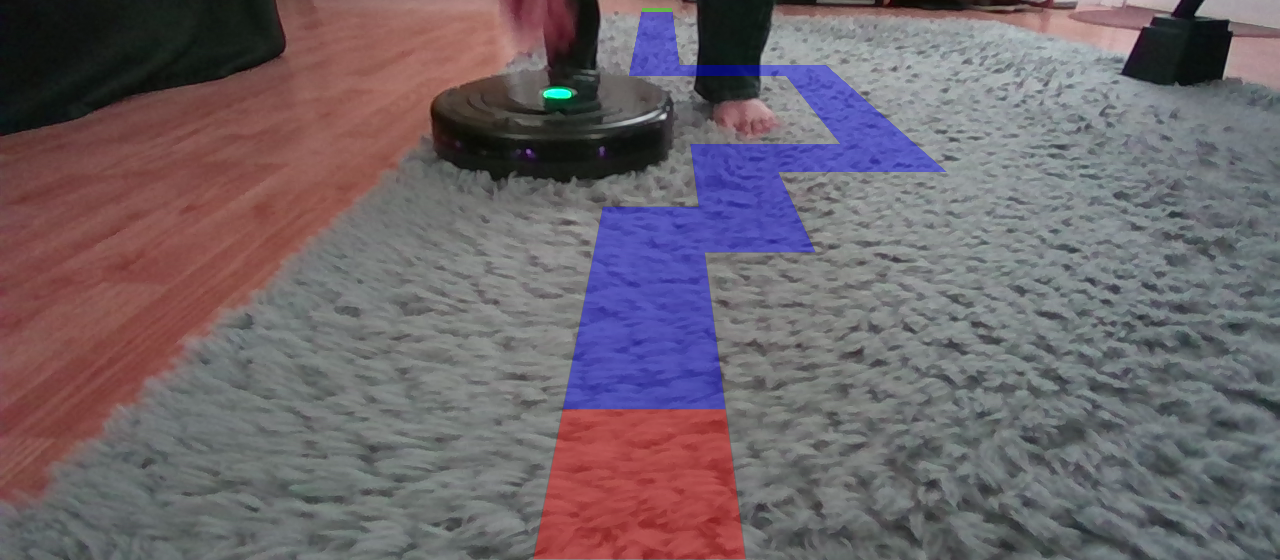}
         \caption{Find Path Behind Obstacles}
         \label{fig:path_behind}
     \end{subfigure}
     \hfill
     \begin{subfigure}[b]{0.3\textwidth}
         \centering
         \includegraphics[width=\textwidth]{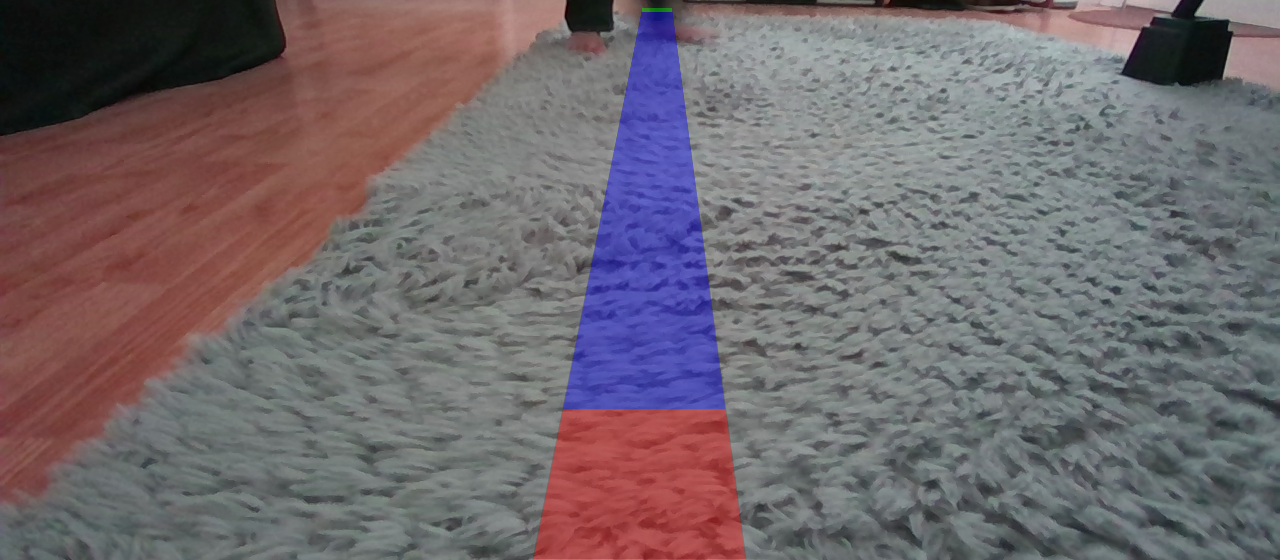}
         \caption{Sub-optimal Path}
         \label{fig:suboptimal_path}
     \end{subfigure}
        \caption{Test Data Results}
        \label{fig:test_results}
\end{figure}

The test data consists of 1200 images, and a qualitative evaluation of the test data is provided in the video (please refer to the code section). Figure \ref{fig:test_results} showcases some of the notable moments captured in the video.

Observing the qualitative and quantitative results, it becomes evident that certain paths contain errors, particularly when the walking robot encounters a blocked path. These errors will be further analyzed and discussed in detail in the Error Analysis section.

\section{Error Analysis}

Our experiments demonstrate that the navigation system exhibits several strengths, yet there are notable limitations that need to be addressed. The issues are exemplified by the blocked paths depicted in Figures \ref{fig:log_1_results}, \ref{fig:log_2_results}, and \ref{fig:test_results}. In Figure \ref{fig:blocked_2}, the semantic segmentation fails to recognize the carpet beyond the obstacles, resulting in a path planning error where the robot attempts to navigate through the highest-cost route. Figure \ref{fig:blocked_1} demonstrates a situation where the vision system identifies a small gap and plans a path through it, despite the practical impossibility for the robot to traverse this path successfully. In Figure \ref{fig:5_obst_2}, a slightly skewed path is taken due to an erratic boundary in the segmentation network's output. Furthermore, Figure \ref{fig:suboptimal_path} illustrates a scenario where an obstacle near the destination causes the search algorithm to choose a sub-optimal path that traverses the obstacle.

The main takeaway from these examples is that the accuracy of the segmentation system directly influences the accuracy of the robot's navigated path. It is worth noting that the segmentation network tends to have lower accuracy at longer distances. However, this limitation can be mitigated to some extent as robots are expected to continuously re-plan their paths while moving forward. Shorter distances allow for error correction, but it is important to recognize that this may lead to sub-optimal paths overall.

Beyond the analysis discussed above, one of the major challenges here is the robot's difficulty in perceiving and planning paths beyond tall obstacles, which hinders its ability to navigate effectively in such scenarios. The take-away is that to tackle the obstacle of tall obstacles, it is worthwhile to consider investigating partially-observable search methods in future research.

\section{Future Work}

Looking beyond the scope of this project, our research will focus on further enhancing the obstacle-avoidance navigation system by investigating advanced path-planning techniques. One such technique is performing search in partially-observable environments to plan paths in scenarios where the environment is not entirely observable to the robot. 

Furthermore, our future work will involve expanding the scope of the research to encompass a wider range of home environments, including stairs and uneven surfaces. Additionally, we will explore techniques to optimize the efficiency of the system in real-time scenarios, taking into consideration factors such as robot stability, dynamic obstacle avoidance, and resource constraints.

\section{Code and Video}

Repository: \href{https://github.com/manglanisagar/vision-search-navigation}{Github Link}

Note: The offline code is currently available for testing purposes, while the online code, which is specifically designed for deployment on the robot, will be released after publication.

Video: \href{https://youtu.be/CTwg6dD-oxI}{Youtube Link}

Dataset Links:

1. Labeled Data: \href{https://drive.google.com/drive/folders/1sxXblBL04injdSfNBE3NMQGg7dtDwJAs?usp=sharing}{Google Drive Link}

2. Unlabeled Data: \href{https://drive.google.com/drive/folders/1xe9N7UEEH2GSFKTQ7Z9DMfBM1-183ovO?usp=sharing}{Google Drive Link}

\bibliographystyle{unsrt}

\end{document}